\documentclass[preprint,review,12pt]{elsarticle}

\usepackage{amssymb}
\usepackage{amsmath}
\usepackage{algorithm}
\usepackage{algorithmic}
\usepackage{booktabs}
\usepackage{multirow}
\usepackage{amsmath}
\usepackage{subcaption}
\usepackage{tabularx}
    \newcolumntype{L}{>{\raggedright\arraybackslash}X}
\usepackage{newfloat}
\usepackage{listings}
\floatstyle{ruled}
\newfloat{listing}{tb}{lst}{}
\floatname{listing}{Listing}

\journal{Pattern Recognition}

\begin{document}

\begin{frontmatter}




\title{A Language-Guided Benchmark for Weakly Supervised Open Vocabulary Semantic Segmentation}


\author[inst1]{Prashant Pandey}

\affiliation[inst1]{organization={Department of Electrical Engineering},
            addressline={Indian Institute of Technology}, 
            city={New Delhi},
            country={India}}

\author[inst2]{Mustafa Chasmai}
\author[inst3]{Monish Natarajan}
\author[inst1]{Brejesh Lall}

\affiliation[inst2]{organization={Department of Computer Science },
            addressline={Indian Institute of Technology}, 
            city={New Delhi},
            country={India}}

\affiliation[inst3]{organization={Department of Computer Science },
            addressline={Indian Institute of Technology}, 
            city={Kharagpur},
            country={India}}

\begin{abstract}
Increasing attention is being diverted to data-efficient problem settings like Open Vocabulary Semantic Segmentation (OVSS) which deals with segmenting an \textit{arbitrary} object that may or may not be seen during training. The closest standard problems related to OVSS are Zero-Shot and Few-Shot Segmentation (ZSS, FSS) and their Cross-dataset variants where zero to few annotations are needed to segment novel classes. The existing FSS and ZSS methods utilize fully supervised pixel-labelled seen classes to segment unseen classes. Pixel-level labels are hard to obtain, and using weak supervision in the form of inexpensive image-level labels is often more practical. To this end, we propose a novel unified weakly supervised OVSS pipeline that can perform ZSS, FSS and Cross-dataset segmentation on novel classes \textit{without using pixel-level labels for either the base (seen) or the novel (unseen) classes in an inductive setting}. We propose \textbf{W}eakly-Supervised \textbf{L}anguage-Guided \textbf{Seg}mentation \textbf{Net}work (WLSegNet), a novel language-guided segmentation pipeline that i) learns generalizable context vectors with batch aggregates (mean) to map class prompts to image features using frozen CLIP (a vision-language model) and ii) decouples weak ZSS/FSS into weak semantic segmentation and Zero-Shot segmentation. The learned context vectors avoid overfitting on seen classes during training and transfer better to novel classes during testing. WLSegNet avoids fine-tuning and the use of external datasets during training. The proposed pipeline beats existing methods for weak generalized Zero-Shot and weak Few-Shot semantic segmentation by 39 and 3 mIOU points respectively on PASCAL VOC and weak Few-Shot semantic segmentation by 5 mIOU points on MS COCO. On a harder setting of 2-way 1-shot weak FSS, WLSegNet beats the baselines by 13 and 22 mIOU points on PASCAL VOC and MS COCO, respectively. Without using dense pixel-level annotations, our results for MS COCO ZSS are comparable to fully supervised ZSS methods. We also benchmark weakly supervised Cross-dataset Segmentation.
\end{abstract}


\begin{highlights}
\item First method to explore multiple and related Open Vocabulary Semantic Segmentation inductive tasks in a weakly supervised setting without using external datasets and fine-tuning
\item First method to handle weakly supervised generalized zero-shot segmentation, zero-shot segmentation and few-shot segmentation with a single training procedure using a frozen vision-language model 
\item Propose a novel and scalable mean instance aware prompt learning that generates highly generalizable prompts, handles domain shift across the datasets and generalizes efficiently to unseen classes
\item The flexible design allows easy modification and optimization of different components as and when required  
\item The proposed method beats existing weakly supervised baselines by large margins while being competitive with pixel-based methods
\end{highlights}


\end{frontmatter}


\section{Introduction}

Since its first arrival, semantic segmentation has gained a lot of attention and remarkable progress has been made in this field. The advent of deep learning led to a new era with methods aiming to surpass human performance. However, most semantic segmentation methods are trained with dense pixel-level annotations and require large numbers of examples for every category or task. 
Humans, in contrast, are able to recognize or at least have some context about new objects without ever seeing them before. Although natural to humans, this task requires a complex understanding of the semantic meaning of a class/category never seen by a learner and a capacity to generalize knowledge gained from seen classes. Most fully supervised methods, although performing comparably with humans on seen objects, struggle in this generalisation to unseen objects. Obtaining large numbers of fully  annotated data for every target class can be extremely expensive and often impractical, so there is a need for models to be able to generalize to unseen classes. 

To bridge the gap between human and artificial learning, increasing attention is being diverted to more challenging and data-efficient settings like Open Vocabulary Semantic Segmentation (OVSS)~\cite{xu2021simple,ma2022open}. In this paper, we focus on closely-related standard OVSS settings like Generalized Zero-Shot Segmentation (GZSS), Zero-Shot Segmentation (ZSS), Few-Shot Segmentation (FSS) and Cross-dataset Segmentation in an inductive setting (unlabeled pixels and novel class names are not observed during training) as opposed to the transductive setting (unlabeled pixels and novel class names may be observed during training). In ZSS~\cite{bucher2019zero, ding2022decoupling, li2022languagedriven, baek2021exploiting,  liu2022open}, a model is provided with a set of base (seen) classes to learn from and then expected to perform well on the novel (unseen) classes it does not have access to. A commonly used setting Generalised ZSS (GZSS) further imposes the expectation that the model in addition to novel classes should retain its performance on base classes as well. Similar to how humans can relate visual understanding of classes with similar-in-meaning names or categories, ZSS methods generalize semantic visual information using the semantic textual information provided by language models. Another slightly relaxed data efficient setting is FSS~\cite{wang2019panet, Mao_Zhang_Wang_Zhang_Xiang_Pan_2022, kang2022integrative, lang2022learning, iqbal2022msanet, 10.1007/978-3-031-16452-1_8,pandey2022robust}, where the model is expected to generalize to unseen classes but is additionally given few support images with annotated unseen target classes. Typical FSS methods demonstrate admirable performance using support samples ranging from one to five examples for every unseen category. 

Besides ZSS and FSS, many other problem settings aim to reduce the burden of large-scale annotations. A particularly interesting approach is Weakly Supervised Segmentation (WSS)~\cite{jiang2022l2g, zhang2020reliability, zhou2022regional, lee2021railroad}, where costly pixel-level annotations for training classes are replaced with relatively inexpensive weak labels  like scribbles and bounding boxes. A particularly challenging setting here is that of image tags, where every image is accompanied by only the information of classes present in it. Without any information allowing the model to localize objects, this setting is perhaps the \textit{hardest} for WSS. 

\begin{figure}[!h]
\centering
\includegraphics[width=0.75\linewidth]{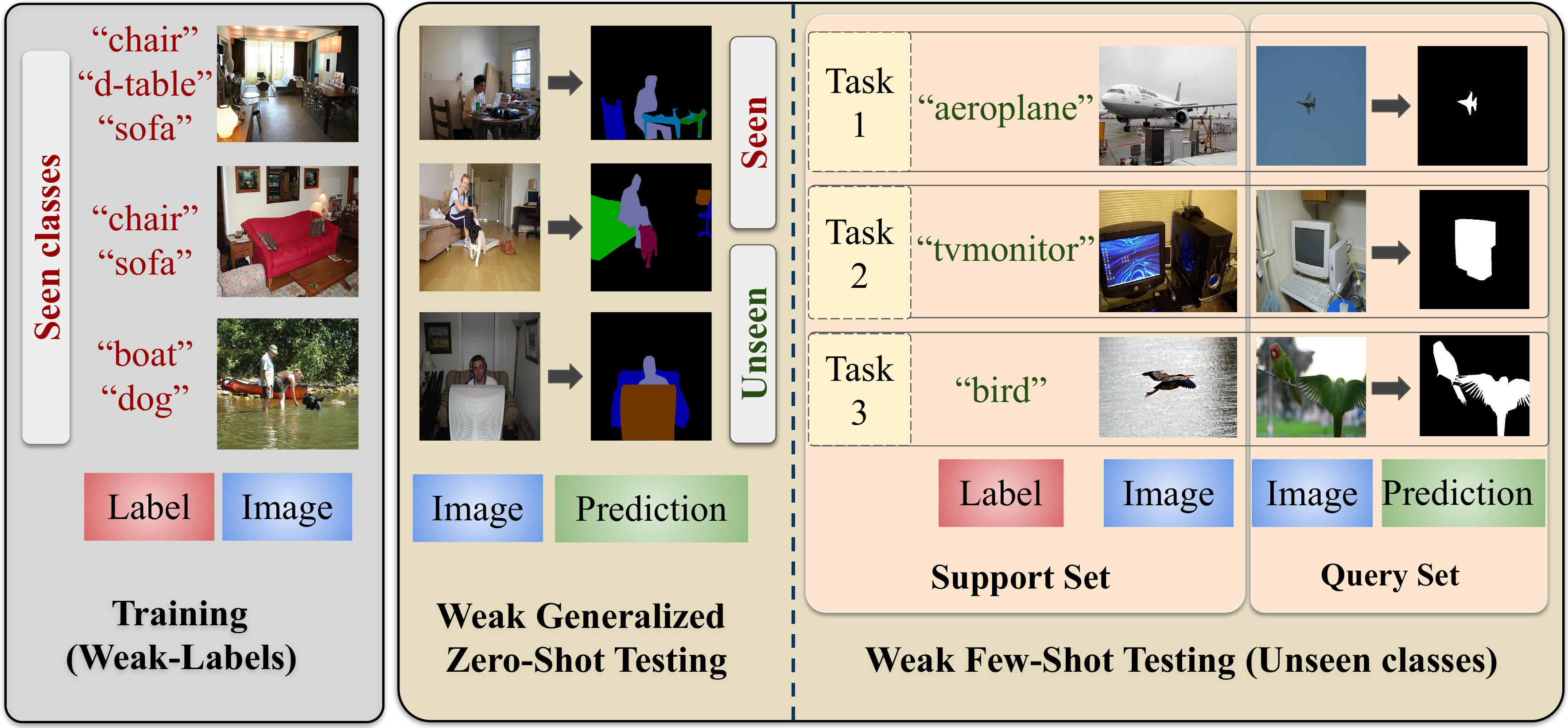}
\caption{An overview of the problem setting explored. A common training procedure is used where a set of seen classes with \textit{only image-level labels} are exposed to the model. The same model is used to evaluate WGZSS and WFSS settings. In WGZSS, the model needs to segment classes ``person" and ``tvmonitor" which it had not seen during training, along with the seen classes. In WFSS, each task has a corresponding target class label in the support and the model segments only that class in the query. For WGZSS, no labels are available during testing while only image-level support labels are present during WFSS testing as opposed to FSS which has pixel-level support labels.}
\label{fig:setting}
\end{figure}

In this paper, we explore the challenging and practical problems of weakly supervised ZSS (WZSS) and weakly supervised FSS (WFSS). With an expectation of generalization to unseen classes and reliance on only weak image-level labels, these settings greatly reduce the annotation cost and assess a method's performance in challenging scenarios commonly faced by humans. A clear rift is evident between existing ZSS and FSS methods, where ZSS methods leverage language model-based learning and try to learn mappings between visual and textual features, while FSS methods tend to employ matching-based approaches that search semantically similar features between support and queries. We argue that when using weak labels, the Few-Shot tasks can also be de-coupled into WSS for learning to segment seen categories and ZSS for generalizing this learning to unseen categories. With this, we propose Weakly-Supervised Language-Guided Segmentation Network (WLSegNet), a unified method that can perform both WZSS and WFSS with a single training procedure. We also benchmark weakly supervised Cross-dataset segmentation setting where we train with weak image-level labels on one dataset (like MS COCO) and test on novel classes of a completely different dataset (like PASCAL VOC).
\begin{figure}[!h]
\centering
\includegraphics[width=0.8\linewidth]{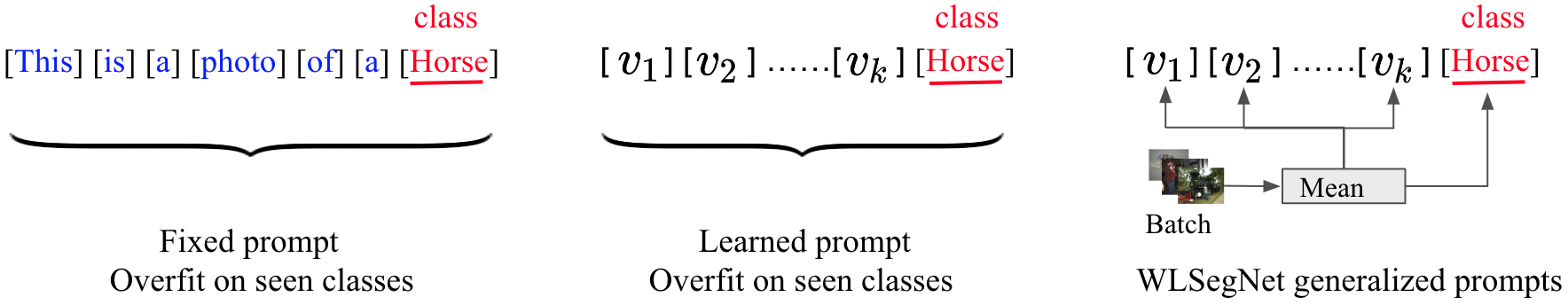}
\caption{An overview of the proposed prompt learning by WLSegNet. The fixed and the learned prompts overfit on seen classes during training while WLSegNet prompts employ mean features from the image batch to make prompts image-aware as opposed to being class-aware thereby avoiding overfitting on seen classes and generalising well on unseen classes. Also, as WLSegNet utilizes a batch of images instead of a single image, this makes the proposed prompt learning more computationally efficient. }
\label{fig:our_prompt_overview}
\end{figure}
We further address limitations like overfitting on seen classes and large computational requirements reported by existing prompt-based learning methods~\cite{zhou2022learning, zhou2022conditional}. We employ batch aggregate (mean) image features to make learnable prompts image-aware, while maintaining low computational requirements. 
The learned prompts avoid overfitting on seen classes and generalise well to novel classes without aid from external datasets or fine-tuning. An overview of our proposed prompt learning approach is shown in Figure~\ref{fig:our_prompt_overview}. In summary, our contributions are fourfold:

\begin{itemize}
    \item We propose a model to perform WZSS, WFSS and Cross-dataset segmentation in a unified manner in an inductive setting. To the best of our knowledge, we are the first to tackle these challenging yet impactful problems together that avoid fine-tuning and the use of external datasets with a frozen vision-language model (CLIP~\cite{radford2021learning}).
    \item We propose an optimal pipeline that decouples the WZSS and WFSS problems into WSS and ZSS. This facilitates the optimization of WSS, mask proposal generation, and vision-language models, separately.
    \item We propose a novel mean instance aware prompt learning method that makes prompts more generalizable and less prone to overfitting while scaling the prompt learning stage to larger batch sizes and improving its computation speed. 
    \item We perform extensive experiments on widely used PASCAL VOC and COCO datasets for WZSS and WFSS, beating previous weakly supervised baselines by large margins and obtaining results competitive with methods using strong supervision. We benchmark Cross-dataset segmentation by training with image-level labels of the COCO dataset and testing on novel classes of PASCAL VOC.
\end{itemize}

\section{Related Work}
\subsection{Zero-Shot Segmentation}
Existing zero-shot methods are broadly generative or discriminative, some of which further incorporate self-training to capture latent features of novel classes. Generative methods include ZS3Net~\cite{bucher2019zero} which trains a generator to generate synthetic features for unseen classes that are used along with real features of seen classes to train the classifier head, and CagNet~\cite{gu2020context} where-in feature generation is guided by contextual-information present in the image. Discriminative methods like SPNET~\cite{xian2019semantic} and LSeg~\cite{li2022languagedriven} map the pixel-level features of an image with the word embeddings of its class obtained from pre-trained word encoders such as word2vec~\cite{mikolov2013efficient} or fastText~\cite{bojanowski2017enriching}. STRICT~\cite{pastore2021closer} employs SPNET as a pseudo-label generator coupled with consistency regularization to improve Zero-Shot performance.  Recent methods such as ZegFormer~\cite{ding2022decoupling} and SimSeg~\cite{xu2021simple} first learn to generate class-agnostic mask proposals using MaskFormer~\cite{cheng2021per}, and then classify proposal regions using knowledge of pre-trained vision-language models such as CLIP~\cite{radford2021learning}. ZegCLIP~\cite{zhou2022zegclip} proposes a one-stage approach that directly extends CLIP’s zero-shot prediction capability from image to pixel level.

\subsection{Weakly Supervised Segmentation}
Weakly Supervised Segmentation methods deal with the practical setting of generating segmentation masks with models trained with weak forms of supervision such as bounding-box~\cite{dai2015boxsup, oh2021background,lee2021bbam}, scribbles~\cite{lin2016scribblesup,vernaza2017learning,liang2022tree}, points~\cite{bearman2016s} and image-level labels. Here we focus on WSS methods using image-level labels only. A commonly used strategy is to train a classifier and then use Class Activation Maps (CAMs) to obtain pixel-level pseudo labels. Some recent methods try to expand the initial seed regions highlighted by CAMs via adversarial erasing~\cite{wei2017object,kumar2017hide,hou2018self}, region growing~\cite{kolesnikov2016seed,huang2018weakly,wang2018weakly}, random-walk~\cite{vernaza2017learning,ahn2018learning,ahn2019weakly} and stochastic inference~\cite{lee2019ficklenet} to name a few. Some methods refine initially coarse attention maps by trying to maximize object region and minimise background coverage. These include EPS~\cite{lee2021railroad} which uses supervision of saliency-maps from off-the-shelf saliency detectors to guide learning and RSCM~\cite{jo2022recurseed} which incorporates high-order feature-correlation and improves masks through seed recursion. CIAN~\cite{fan2020cian} propagates pixel-wise affinity to pixels in the neighbourhood. RCA~\cite{zhou2022regional} maintains a memory bank for storing object features across the images in the dataset which serves as a support during pseudo-label generation. L2G~\cite{jiang2022l2g} transfers features learned by local region-wise classifiers to a global classification network, thus capturing greater detail and obtaining higher-quality attention maps.

\subsection{Semantic Embeddings and Language Models}
Transfer of knowledge from seen to unseen classes requires auxiliary information.
Such information can be provided by the semantic embeddings of class names obtained from word encoders like word2vec~\cite{mikolov2013efficient} and fastText~\cite{bojanowski2017enriching} which are trained on large-scale word datasets without human annotation, with the help of simple axioms like the one that says that words occurring often in similar contexts have closer feature representations. More recently transformer-based vision-language models such as CLIP~\cite{radford2021learning} and ALIGN~\cite{jia2021scaling} have been pre-trained on large-scale image-text pairs from the web in a contrastive manner for zero-shot classification. The key idea in retrieving features in CLIP is to pass a sentence containing a class name along with context information which may be a predefined prompt-template~\cite{ding2022decoupling} or learned~\cite{zhou2022learning,zhou2022conditional}. ~\cite{zhou2022learning} shows that dataset-specific context information in prompt templates improves Zero-Shot classification accuracy using CLIP. In our work, we adopt CLIP and propose a novel prompt-learning technique that incorporates instance-specific context using a batch mean of input features in addition to dataset-specific context learning. A parallel line of works~\cite{ghiasi2022scaling, liang2022open,mukhoti2022open,luo2022segclip,xu2023learning,ren2023viewco} use image-level labels/captions, pre-train/fine-tune the vision-language or language pre-training models (like ALIGN, CLIP, BERT~\cite{devlin2018bert}), which
require large-scale external datasets or perform transductive segmentation~\cite{zhou2022extract}. Fusioner~\cite{ma2022open}, a cross-modality fusion module, explicitly bridges a variety of self-supervised pre-trained visual/language models for open-vocabulary semantic segmentation.

\subsection{Zero and Few-shot Segmentation with Weak Supervision}
Very few works have explored the practical setting of Zero and Few-Shot segmentation with weak annotations. \cite{lee2022pixel} follows a meta-learning approach for few-shot segmentation. For a support image and a given weak label, it generates CAMs for a set of seen classes using a pre-trained network, then performs a weighted summation of these with the weights proportional to similarities of the textual features obtained by word2vec. Similarly, \cite{shen2022dual} first proposed the setting of WZSS using only image labels for seen classes as supervision. Another line of work is open-world segmentation \cite{liu2022open} where models are trained using large-scale image captioning datasets without a need for dense pixel annotations. We take inspiration from previous works and propose a novel pipeline that unifies both Zero-Shot and Few-Shot segmentation using only weak labels as supervision.

\section{Methodology}
\label{sec:methodology}
\subsection{Problem Setting}
The task of WFSS includes train $\mathcal{D}_{\mathrm{train}}$ and test $\mathcal{D}^{F}_{\mathrm{test}}$ weakly labelled datasets having non-overlapping class sets. 
The test dataset $\mathcal{D}^{F}_{\mathrm{test}}$ consists of a set of episodes with each episode containing $N$-way $K$-shot tasks with support and query sets.
The support set $\mathcal{S}_i$ has $K$ image $(\mathcal{I}_{\mathcal{S}})$ and image-level label $(L_{\mathcal{S}})$ pairs with a total of $N$ semantic classes  i.e. $\mathcal{S}_i =\{(\mathcal{I}_{\mathcal{S}}^{k},L_{\mathcal{S}}^{k})\}$ where  $L^{k}_{\mathcal{S}}$ is the ground-truth \textit{image tag} for $k$-th shot, 
and $k=1,2, ..., K$.
The query set $\mathcal{Q}_i$ has $N_{\mathcal{Q}}$ $\mathcal{\mathrm{images}~(I_{Q})}$. The objective in each test episode $i$ is to obtain high-quality segmentation predictions for the query set $\mathcal{Q}_i$, relying on the weakly labelled support set $\mathcal{S}_i$, both of whose classes are never seen during training. 
The training dataset $\mathcal{D}_{\mathrm{train}}$ consists of a set of images and their corresponding image tags (weak labels). A common approach in FSS is to break down $\mathcal{D}_{\mathrm{train}}$ into episodes having support and query sets and then train episodically using metric learning. 
However, there is no restriction on the use of $\mathcal{D}_{\mathrm{train}}$, i.e. a method may instead decide to use the images of support and query sets in a non-episodic way. WZSS is logically an extension of WFSS in that it is simply an N-way 0-shot WFSS task. However, it differs in some critical aspects in its formulation. WZSS consists of $\mathcal{D}_{\mathrm{train}}$ and $\mathcal{D}^{Z}_{\mathrm{test}}$, where the training dataset is identical to that in WFSS but the testing dataset $\mathcal{D}^{Z}_{\mathrm{test}}$ simply consists of a set of images and the set of distinct classes $C_{test}$. Let the set of all distinct classes present in the training dataset $\mathcal{D}_{\mathrm{train}}$ be $C_{train}$. Depending on the nature of $C_{train}$ and $C_{test}$, two different settings are possible. The first, default WZSS setting is when $C_{train} \cap C_{test} = \phi$, i.e. the classes during testing are disjoint from the classes seen during training. The second setting of generalised Zero-Shot (WGZSS) holds when $C_{train} \subset C_{test}$, i.e. the test dataset contains both seen and unseen classes. 


\begin{figure*}[]
\centering
\includegraphics[width=0.75\linewidth]{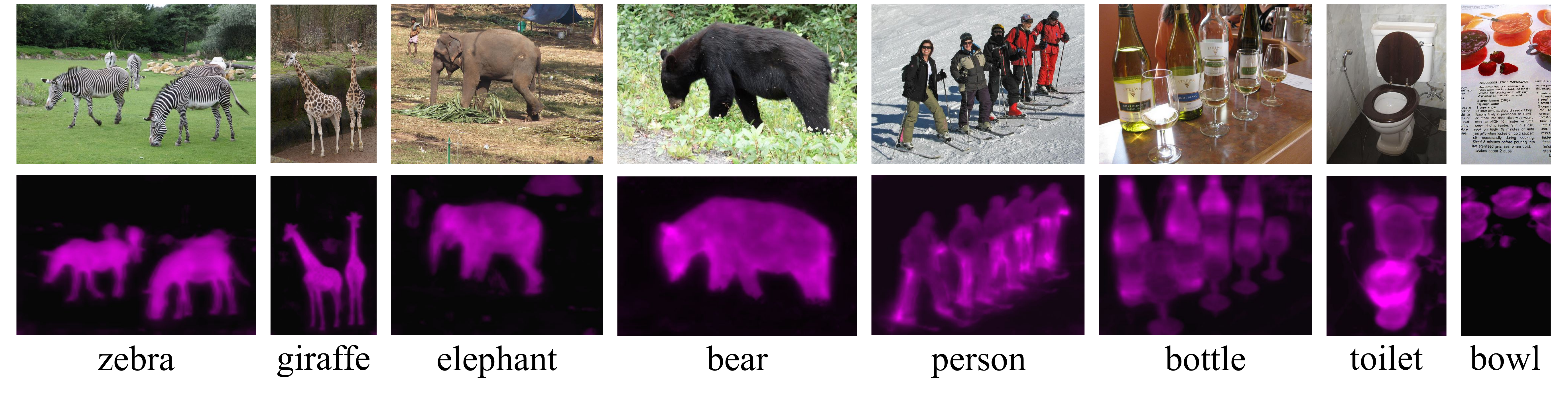}
\caption{Weakly supervised segmentation using image-level labels.}
\label{fig:weak}
\end{figure*}

\subsection{Pseudo-label Generation Module}
We adopt L2G~\cite{jiang2022l2g}, a weakly supervised semantic segmentation method for this task. Traditional CAMs obtained from classification networks tend to only highlight the discriminative regions of an object making it unsuitable for semantic segmentation. L2G transfers the knowledge learnt by multiple local regional classification networks to a single global classification network thus expanding and improving the region of focus in  attention maps. 
In the method, a multi-target classification network is trained on the set of seen classes $C_{train}$. For a given input image $\mathcal{I}$, the corresponding pseudo-segmentation mask $\mathcal{M}_{psuedo}$ is obtained from CAMs produced by the network. We denote the fully trained model as Pseudo-label Generation (PLG) Module. The masks generated by the PLG are then used to train the Class-Agnostic Mask Generation (CAMG) module. A few samples of the attention maps learnt can be seen in Figure~\ref{fig:weak}.

While we use L2G~\cite{jiang2022l2g} as our pseudo-label generator, we would like to point out that it can easily be replaced by another WSS method without changing the overall architecture much due to the decoupled design. 
Thus, rather than being constrained by L2G~\cite{jiang2022l2g}, we can leverage advances in the field of WSS to further improve the performance of WLSegNet. We do such an experiment and observe that with RCA~\cite{zhou2022regional} as the pseudo-label generator, the performance does not change drastically. 

\subsection{Class-Agnostic Mask Generation (CAMG)}
The task of Zero-Shot segmentation is broken down into class-agnostic aggregation of pixels through the CAMG module followed by CLIP classification of aggregated regions (segments). Similar to past works, we adopt MaskFormer~\cite{cheng2021per}, a segmentation model that generates mask proposals for different objects present in the image irrespective of the class of any object. Specifically, for a given image $\mathcal{I}$ fed to the CAMG module, a set $\mathcal{M}$ of $n$ class-agnostic binary mask proposals are generated, such that $\mathcal{M} = \{m_1,m_2, \ldots, m_n\}$. During training, only the pseudo-labels obtained from PLG are used as supervision in MaskFormer’s Mask Loss. 
For each mask or segment proposal $m \in \mathcal{M}$, we create a corresponding input proposal $i$ by multiplying input image $\mathcal{I}$ with $m$ to zero out the background in corresponding segments. The input proposals $\mathcal{I}_p = \{i_1, i_2, \ldots, i_n$\} thus obtained are then passed to CLIP for segment classification. 

Instead of MaskFormer~\cite{cheng2021per}, the CAMG module can use other methods like GPB-UCM~\cite{arbelaez2010contour} and Selective Search~\cite{uijlings2013selective} as well. Since~\cite{xu2021simple} showed superior performance of MaskFormer, we perform all our experiments with it.

\subsection{CLIP language model}
For Zero-Shot classification of an image using CLIP, we first feed the CLIP text encoder with a corresponding text label in the form of a natural sentence for each class present. This can be a simple prompt such as “photo of a \{class\}” where \{class\} is replaced by the appropriate class name. However, such simple fixed prompts fail to capture context information present in the image and this affects the performance of the downstream task of segment classification. 

\begin{figure}[]
\centering
\includegraphics[width=.75\linewidth]{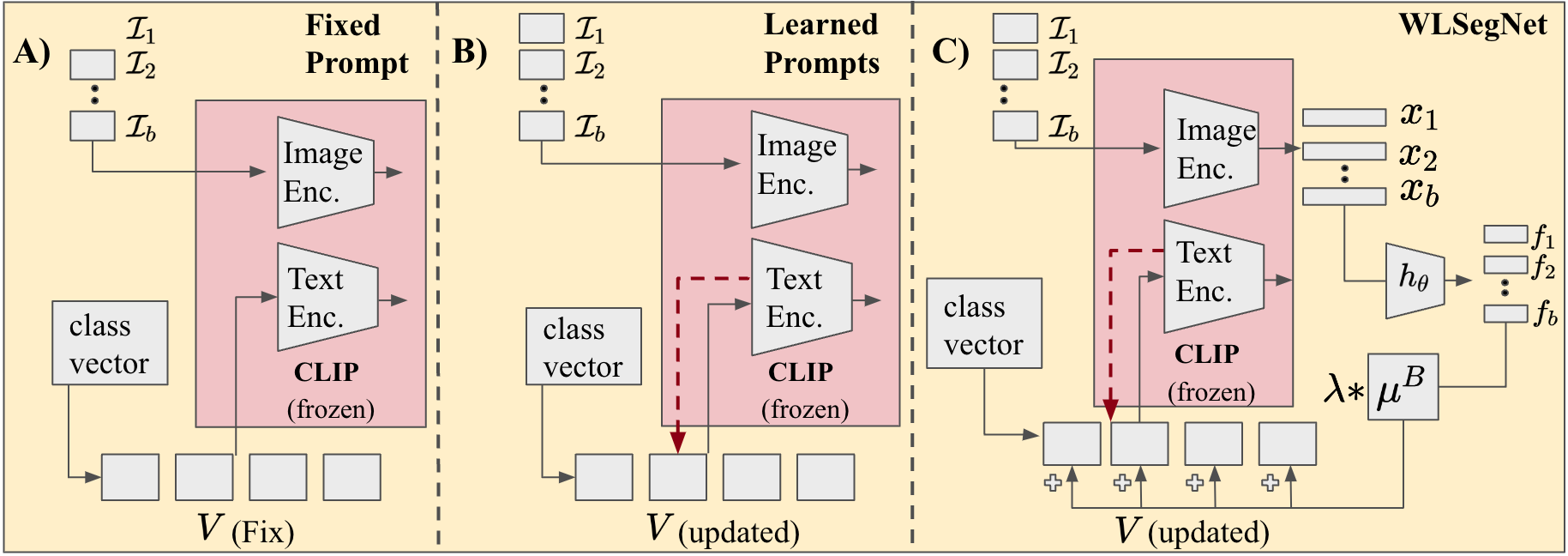}
\caption{Comparison of the proposed prompt learning strategy in WLSegNet with prior works. A) Fixed prompts~\cite{ding2022decoupling}, B) Learned prompts~\cite{zhou2022learning} with dataset-specific context vectors, C) WLSegNet (Ours): Batch mean instance features ($\mu^B$) incorporated into learned context vectors. $\lambda$ controls the extent of $\mu^B$ added to context vector $V$. Red-dotted arrow denotes backpropagation to update the context vector. During prompt learning, CLIP encoders are frozen.
}
\label{fig:prompt}
\end{figure}

\subsubsection{Prompt Learning}
To overcome the limitations of fixed prompts, many recent learned prompt works \cite{zhou2022learning, deng2022learning} propose to incorporate dataset-specific context information by using learnable prompt context vectors that can be catered to work best for each particular class. These learned prompts are biased towards seen classes. \cite{zhou2022conditional} further improved upon this by using an additional input-conditional token, making the prompts less sensitive to class shift and thus, more generalizable to unseen classes. However, they report increased computation and resulting restrictions on the batch size. We overcome these limitations and propose a mean instance aware prompt learning strategy to learn better and more generalizable prompts. A brief overview of the different prompt strategies can be seen in Figure~\ref{fig:prompt} for a better comparison. 

Specifically, our prompt learning approach learns a context vector $V$ to capture context information of the dataset. $V$ is constructed such that it contains $k$ prompt tokens each of dimension $d$, expressed by $V = [v]_1[v]_2...[v]_k$. Class prompt proposal $V_c$ is then obtained by concatenating the context vector $V$ with the class embedding of class $c$ such that $V_c = [v]_1[v]_2...[v]_k[w_c]$ where $w_c$ represents the class embedding. For a given image $\mathcal{I}$ in input batch $B$ of size $b$, let $x$ be the image embedding obtained from the pre-trained CLIP Image Encoder. Image embedding $x$ is passed through a shallow neural network represented by $h_\theta(.)$ to obtain the instance-wise features $f = h_\theta(x)$. We then obtain the mean batch feature prototype $\mu^B$ as shown in Eq~(\ref{eq:mean}).

\begin{equation}
\label{eq:mean}
\mu^{B}=\frac{1}{b} \sum f_{i}, i \in\{1,2, \ldots, b\}    
\end{equation}
Finally, for a given batch the mean instance aware class prompt $V_c^B$ is obtained as shown in Eq~(\ref{eq:lambda}), which is fed to the pre-trained CLIP text encoder $g(.)$ for Zero-Shot classification. 
\begin{equation} \label{eq:lambda}
V_c^B = V_c + \lambda * G^B
\end{equation}
$G^B$ is $d \times (k+1)$ matrix with each column as $\mu^{B}$ repeated $(k+1)$ times. The extent of $\mu^B$ added to $V_c$ is controlled by a hyperparameter $\lambda$.
The class prediction probability for the given image is computed as shown in Eq~(\ref{eq:preduction}), where $t_c^B$ represents the text embedding of $V_c^B$ obtained from the CLIP text encoder, $C$ is the number of classes, $sim$ is cosine similarity and $\tau$ is the temperature coefficient.
\begin{equation}\label{eq:preduction}
    p(y = c \mid x)=\frac{\exp \left(sim\left(x, t_c^B\right) / \tau\right)}{\sum_{i=1}^{C} \exp \left(sim\left(x,t_i^B)  / \tau\right)\right.}
\end{equation}
The class predictions, combined with mask proposals are then aggregated to obtain semantic segmentation output. This provides more generalizable prompts compared to existing methods shown in Figure~\ref{fig:prompt}. WLSegNet facilitates scaling of prompt learning to larger batch size and thereby it is computationally faster than \cite{zhou2022conditional} while also being less prone to overfitting on seen classes.

\begin{figure*}[]
\centering
\includegraphics[width=1\linewidth]{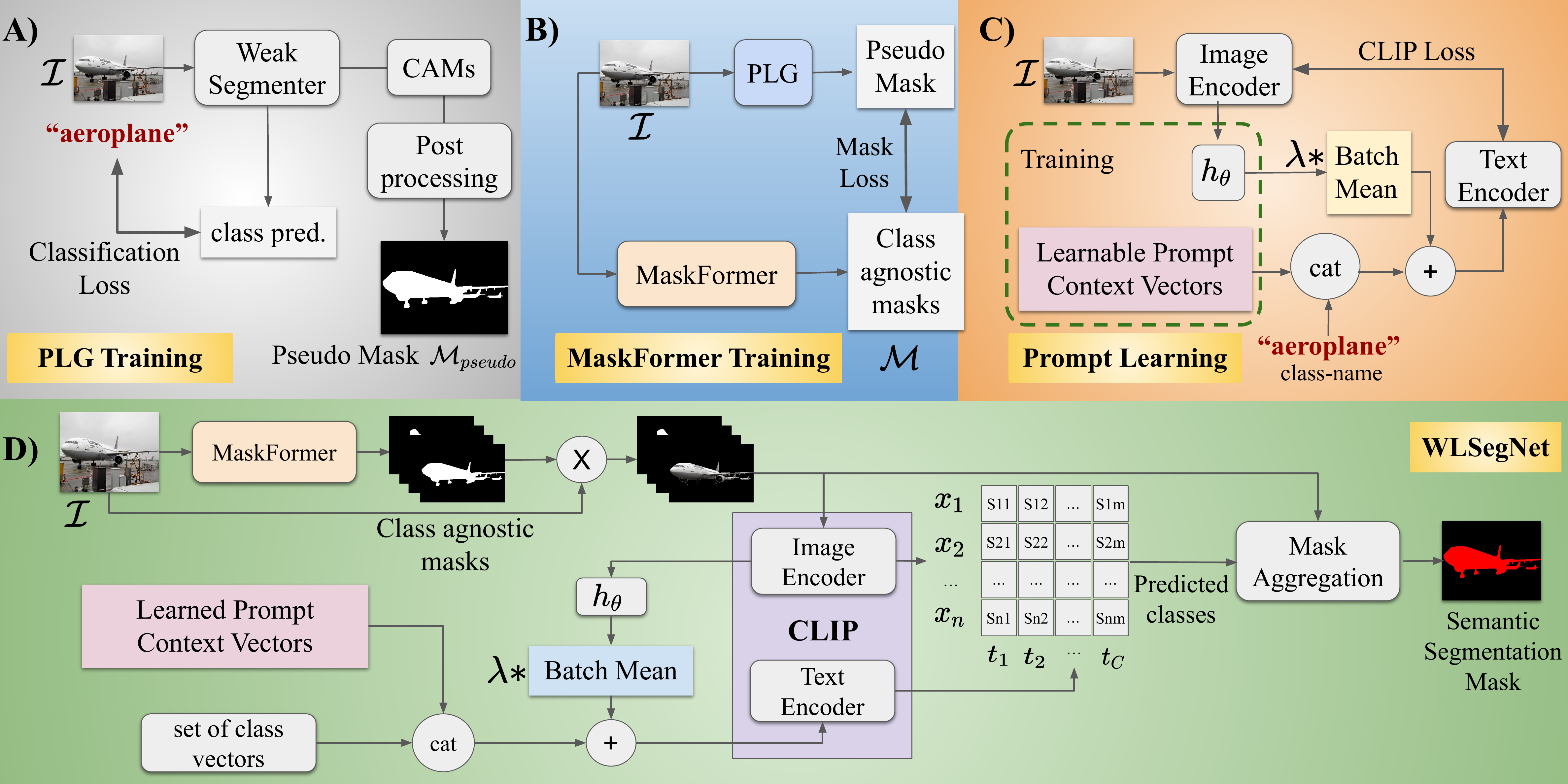}
\caption{WLSegNet consists of a \textbf{(A)} Pseudo-Label Generation (PLG) module that uses image-level supervision to generate pseudo-segmentation mask $\mathcal{M}_{psuedo}$ for input image $\mathcal{I}$ among seen classes in $C_{train}$, \textbf{(C)} A prompt learning module that incorporates batch mean instance features with learnable context vectors to generate instance aware prompts and \textbf{(B, D)} where $\mathcal{M}_{psuedo}$ generated by the PLG module governs training of the Class-Agnostic Mask Generation (CAMG) module which produces a set of class-agnostic mask proposals $\mathcal{M}$. The mask proposals in $\mathcal{M}$ are multiplied with $\mathcal{I}$ and passed to CLIP (frozen) for segment classification among $C_{test}$ classes, along with the learned class prompts. The segmented masks and corresponding class predictions thus obtained are aggregated to produce the final segmentation mask. $\mathrm{S}_{ij}$ is the cosine similarity between image embedding $x_i$ corresponding to $i^\mathrm{th}$ mask proposal and text embedding $t_j$ corresponding to class $j$.}
\label{fig:methodology}
\end{figure*}

\subsection{Mask Aggregation}

For every class $c$, the mean instance aware class prompt $V_c^B$ is passed to the CLIP text encoder to obtain semantic embedding $t_c^B$, which is used as the weights for classifying the segment embeddings of $\mathcal{M}$ obtained from the CLIP Image Encoder. Note that some regions in the different proposals may overlap. Thus, the segmentation map $\mathcal{Z}$ is obtained by aggregating the different classified proposals as shown in Eq~(\ref{eq:aggregation}).
\begin{equation}\label{eq:aggregation}
    \mathcal{Z}_{j}(q)=\frac{\sum_i m_{i}^{p}(q) C_{i}^{p}(j)}{\sum_k \sum_i m_{i}^{p}(q) C_{i}^{p}(k)}
\end{equation}
Here, $m_i^p(q)$ represents the predicted probability of pixel $q$ belonging to the $i$-th mask proposal $m_i$, and ${C}_i^p(j)$ is the predicted probability of mask proposals $m_i$  belonging to $j$-th category. This pixel-wise class probability $\mathcal{Z}_{j}(q)$ is the final semantic segmentation output.

\subsection{Weak Zero and Few-Shot Inference}
Our method unifies both Zero-Shot and Few-Shot segmentation objectives with common training but different strategies for inference. During (generalized) Zero-Shot testing, for each input image the model segments pixels into seen and unseen classes. The prompts used by CLIP are kept the same for all images, containing one prompt for each class. On the other hand, in the Few-Shot evaluation, the only classes predicted for a particular query are those present in the weak label of the support of a particular task. Thus, the set of prompts used by CLIP varies across different tasks. This subtle difference can also be seen in Figure~\ref{fig:setting}. Additionally for Few-Shot inference, we utilize saliency maps generated from off-the-shelf saliency detectors, as done in prior WSS works like EPS~\cite{lee2021railroad}. The saliency maps help refine the prediction of the difficult-to-describe background class while maintaining predictions of the foreground. 

An overview of the complete procedure can be seen in Figure~\ref{fig:methodology}. Note how PLG training is completely decoupled from the rest of the method. This structure has certain desirable qualities. Since the tasks of generating good pseudo labels for seen classes and generalization to unseen classes are completely decoupled, they can be developed or optimized independently. 
Besides, the unified approach greatly reduces the training cost since a single training is sufficient for evaluation on WGZSS, WZSS, WFSS and Cross-dataset settings.

\section{Implementation Details}

We use one Nvidia A100 GPU to conduct our experiments on PASCAL VOC and 6 1080 GPUs for our experiments on MS COCO. For pixel pseudo labelling, we work with the ResNet38 backbone commonly used in WSS literature. For mask proposals, we use a ResNet50 backbone for COCO and ResNet101 backbone for PASCAL VOC, while for the CLIP language model, we use a ViT-B/16 backbone. CLIP remains frozen during our training, and we initialise it with pre-trained weights trained on publicly available image-caption data. All COCO experiments were trained on 6 GPUs with a batch size of 32 while PASCAL VOC experiments had a batch size of 16. We choose the value 0.01 for the tradeoff hyperparameter $\lambda$ and 0.01 for the temperature $\tau$. For $h_\theta(.)$, we use 2 fully connected layers separated by a ReLU in between. Embedding dimensions of both image and text are 512, and the size of dense layers is chosen to ensure that dimensions match. Other relevant hyperparameters are kept the same as previous works~\cite{jiang2022l2g, xu2021simple}.  All our implementations can be found here: https://github.com/mustafa1728/WLSegNet.

\section{Experiments and Results}

\subsection{Datasets}


\begin{table}[!t]
    \centering
    \caption{GZSS results on PASCAL VOC 2012. Seen classes: 1-15, Unseen classes: 16-20.}
    \scalebox{0.6}{
    \begin{tabular}{clcccc}
    \toprule
        \multirow{2}{*}{Supervision} & \multirow{2}{*}{Method} & \multirow{2}{*}{Venue} & \multicolumn{3}{c}{ mIOU }\\
        & & & seen & unseen & harmonic \\
    \midrule
        \multirow{9}{*}{Pixel Labels} & ZS3Net~\cite{bucher2019zero} & NeurIPS'19 & 78.0 & 21.2 & 33.3\\
         & SPNet~\cite{xian2019semantic} & CVPR'19 & 77.8 & 25.8 & 38.8\\
         & CaGNet~\cite{gu2020context} & ACM MM'20 & 78.6 & 30.3 & 43.7\\
         & SIGN~\cite{cheng2021sign} & CVPR'21 & 83.5 & 41.3 & 55.3\\
         & STRICT~\cite{pastore2021closer} & CVPRW'21 & 82.7 & 35.6 & 49.8\\
         & Joint~\cite{baek2021exploiting} & ICCV'21 & 77.7 & 32.5 & 45.9\\
         & ZegFormer~\cite{ding2022decoupling} & CVPR'22 & 86.4 & 63.6 & 73.3\\
         & SimSeg~\cite{xu2021simple} & ECCV'22 & 79.2 & \textbf{78.1} & 79.3\\
          & ZegCLIP~\cite{zhou2022zegclip} & - & \textbf{91.9} & 77.8 & \textbf{84.3}\\
    \midrule 
        \multirow{3}{*}{Image Labels} & ViL-Seg~\cite{liu2022open} & ECCV'22 & - & 37.3 & -\\
        & DSG~\cite{shen2022dual} & Multimedia'22 & 57.7 & 22.0 & 31.8\\
         & Ours (WLSegNet) & - & \textbf{86.5} & \textbf{59.9} & \textbf{70.8}\\
    \bottomrule
    \end{tabular}
    }
    
    \label{tab:wgzss_pascal}
\end{table}

\begin{table}[t]
\centering
\caption{ZSS results on Pascal-5$^i$. The second best results are underlined.}
\scalebox{0.6}{
    \centering
    
    \begin{tabular}{clcccccc}
    \toprule
        \multirow{2}{*}{Supervision} & \multirow{2}{*}{Method} & \multirow{2}{*}{Venue} & \multicolumn{4}{c}{Fold unseen mIOU} & \multirow{2}{*}{mean}\\
        & & & 0 & 1 & 2 & 3 & \\
    \midrule
        \multirow{6}{*}{Pixel Labels} & SPNet~\cite{xian2019semantic} & CVPR'19 & 23.8 & 17.0 & 14.1 & 18.3 & 18.3\\
        & ZS3Net~\cite{bucher2019zero} & NeurIPS'19 & 40.8 & 39.4 & 39.3 & 33.6 & 38.3\\
         & LSeg~\cite{li2022languagedriven} (ResNet101) & ICLR'22 & \underline{52.8} & 53.8 & \textbf{44.4} & 38.5 & 47.4\\
         & LSeg~\cite{li2022languagedriven} (ViT-L/16) & ICLR'22 & \textbf{61.3} & \textbf{63.6} & \underline{43.1} & \underline{41.0} & \textbf{52.3}\\
         & Fusioner~\cite{ma2022open} (ResNet101) & BMVC'22 & 46.8 & \underline{56.0} & 42.2 & 40.7 & 46.4\\
    \midrule
        \multirow{1}{*}{Image Labels} & Ours (WLSegNet) & - & 47.5 & 47.3 & 39.7 & \textbf{58.5} & \underline{48.2} \\
    \bottomrule
    \end{tabular}
    }
    \label{tab:wzss_pascal}
\end{table}
    

\begin{table}[t]
\centering
\caption{ZSS results on COCO-20$^i$. The second best results are underlined.}
\scalebox{0.6}{
    \centering
    
    \begin{tabular}{clcccccc}
    \toprule
        \multirow{2}{*}{Supervision} & \multirow{2}{*}{Method} & \multirow{2}{*}{Venue} & \multicolumn{4}{c}{Fold unseen mIOU} & \multirow{2}{*}{mean}\\
        & & & 0 & 1 & 2 & 3 & \\
    \midrule
        \multirow{3}{*}{Pixel Labels} & ZS3Net~\cite{bucher2019zero} & NeurIPS'19 & 18.8 & 20.1 & 24.8 & 20.5 & 21.1\\
         & LSeg~\cite{li2022languagedriven} (ResNet101) & ICLR'22 & 22.1 & 25.1 & 24.9 & 21.5 & 23.4\\
         & LSeg~\cite{li2022languagedriven} (ViT-L/16) & ICLR'22 & \textbf{28.1}  & \underline{27.5} & \textbf{30.0} & \underline{23.2} & \underline{27.2}\\
         & Fusioner~\cite{ma2022open} (ResNet101) & BMVC'22 & \underline{26.7} & \textbf{34.1} & 26.3 & \textbf{23.4} & \textbf{27.6}\\
    \midrule
        \multirow{1}{*}{Image Labels} & Ours (WLSegNet) & - & 8.33 & 15.3 & \underline{26.8} & 16.0 & 16.6 \\
    \bottomrule
    \end{tabular}
    }
    \label{tab:wzss_coco}
\end{table}

We perform experiments with PASCAL VOC and MS COCO datasets, keeping settings similar to previous works. Our evaluation metrics closely follow the conventions used in \cite{lee2022pixel}.

\subsubsection{PASCAL VOC 2012} 
This dataset consists of 11185 training images and 1449 validation images, with a total of 20 semantic classes. To compare with WFSS and WZSS methods, we use the Pascal-$5^i$ splits commonly used in FSS literature. From the dataset of 20 classes, 4 folds are created by splitting the classes such that in each fold, 15 classes are seen during training and 5 are reserved as novel classes for testing. Most previous works employing generalised ZSS use a fixed set of seen and unseen classes (classes 1-15 seen, 16-20 unseen). We take the same split while comparing these WGZSS and GZSS methods. While a model would predict all classes (seen and unseen) in these generalised settings, ZSS and WZSS on a particular fold of Pascal-$5^i$ involve the prediction of only unseen classes. This differs from the unseen-mIOU in GZSS primarily in the number of classes being predicted at a time, and one can expect similar performances in both. The model is trained on the training set images with seen classes retained and unseen classes ignored. Evaluation is on the validation images with the novel (and also seen for WGZSS) classes retained. 

\subsubsection{MS COCO 2014} 
This dataset consists of a total of 82081 training images and 40137 validation images, with a total of 80 semantic classes. Similar to PASCAL VOC, we employ the COCO-$20^i$ splits used in literature with 4 folds created by splitting classes into 60 seen and 20 unseen classes. 





\subsection{Results and Discussion}
We have selected pixel-level and weakly supervised Zero and Few-shot segmentation methods as our baselines. Also, we compare with Open Vocabulary Segmentation methods like SimSeg~\cite{xu2021simple} and Fusioner~\cite{ma2022open} that perform segmentation in an inductive setting without pre-training/fine-tuning the vision-language/language models with large-scale external datasets.
\subsubsection{Weakly Supervised Zero-Shot Segmentation}
The performance of our approach in GWZSS and WZSS settings can be seen in Table~\ref{tab:wgzss_pascal}, Table~\ref{tab:wzss_pascal} and Table~\ref{tab:wzss_coco}. This domain is highly under-explored and we do not have baselines strictly following the same setting. Nonetheless, we compare WLSegNet with other strongly supervised methods. It can be seen in Table~\ref{tab:wgzss_pascal} that WLSegNet, using only image labels, beats 6 of the 9 baselines that use dense pixel labels. DSG~\cite{shen2022dual} works  on the same WZSS setting we explore and our method outperforms it by 28.8, 37 and 39 mIOU points for the seen, unseen and harmonic IOU measures. DSG~\cite{shen2022dual} does not report results on COCO or COCO-stuff, so we do not have a comparable baseline for this dataset. Nevertheless, our results on PASCAL VOC and COCO are comparable with strongly supervised baselines, as can be seen in Table~\ref{tab:wzss_pascal} and Table~\ref{tab:wzss_coco}, respectively.

\begin{table}[!t]
\centering
\caption{1-way 1-shot FSS results on Pascal-5$^i$.}
\scalebox{0.6}{
    \begin{tabular}{clcccccc}
    \toprule
        \multirow{2}{*}{Supervision} & \multirow{2}{*}{Method} & \multirow{2}{*}{Venue} & \multicolumn{4}{c}{Fold mIOU} & \multirow{2}{*}{mean}\\
        & & & 0 & 1 & 2 & 3 & \\
    \midrule
        \multirow{6}{*}{Pixel Labels} & PANet~\cite{wang2019panet} & ICCV'19 & 42.3 & 58.0 & 51.1 & 41.2 & 48.1\\
         & CyCTR~\cite{zhang2021few} & NeurIPS'21 & 67.2 & 71.1 & 57.6 & 59 & 63.7\\
         & DPNet~\cite{Mao_Zhang_Wang_Zhang_Xiang_Pan_2022} & AAAI'22 & 60.7 & 69.5 & 62.8 & 58.0 & 62.7\\
         & ASNet~\cite{kang2022integrative} & CVPR'22 & 68.9 & 71.7 & 61.1 & 62.7 & 66.1\\
         & BAM~\cite{lang2022learning} & CVPR'22 & 68.97 & 73.59 & \textbf{67.55} & 61.13 & 67.81\\
         & MSANet~\cite{iqbal2022msanet} & - & \textbf{70.8} & \textbf{75.2} & 67.25 & \textbf{64.28} & \textbf{69.13}\\
    \midrule
        \multirow{1}{*}{B-Boxes} & PANet~\cite{wang2019panet} & ICCV'19 & - & - & - & - & 45.1 \\
    \midrule
        \multirow{1}{*}{Scribbles} & PANet~\cite{wang2019panet} & ICCV'19 & - & - & - & - & 44.8 \\
    \midrule
        \multirow{6}{*}{Image Labels} & PANet~\cite{wang2019panet} & ICCV'19 & 25.7 & 33.4 & 28.8 & 20.7 & 27.1 \\
        & AMP~\cite{siam2019amp} & ICCV'19 & 10.6 & 14.1 & 7.6 & 10.9 & 10.8 \\
        & PFENet~\cite{tian2020prior} & TPAMI'20 & 33.4 & 42.5 & 43.6 & 39.9 & 39.9 \\
        & Pix-MetaNet~\cite{lee2022pixel} & WACV'22 & 36.5 & \textbf{51.7} & 45.9 & 35.6 & 42.4 \\
         & Ours (WLSegNet) ResNet50 & -  & 41.7 & 51.3 & 42.2 & \textbf{41.8	} & 44.2 \\
         & Ours (WLSegNet) ResNet101 & -  & \textbf{45.9} & 46.9 & \textbf{47.2} & 41.5 & \textbf{45.4} \\
    \bottomrule
    \end{tabular}
    }
    
    \label{tab:wfss_pascal}
\end{table}


\begin{table}[t]
    \centering
    \caption{1-way 1-shot FSS results on COCO-20$^i$.}
    \scalebox{0.6}{
    \begin{tabular}{clcccccc}
    \toprule
        \multirow{2}{*}{Supervision} & \multirow{2}{*}{Method} & \multirow{2}{*}{Venue} & \multicolumn{4}{c}{Fold mIOU} & \multirow{2}{*}{mean}\\
        & & & 0 & 1 & 2 & 3 & \\
    \midrule
        \multirow{6}{*}{Pixel Labels} & PANet~\cite{wang2019panet} & ICCV'19 & - & - & - & - & 20.9\\
         & CyCTR~\cite{zhang2021few} & NeurIPS'21 & 38.9 & 43.0 & 39.6 & 39.8 & 40.3\\
         & DPNet~\cite{Mao_Zhang_Wang_Zhang_Xiang_Pan_2022} & AAAI'22 & - & - & - & - & 37.2 \\
         & ASNet~\cite{kang2022integrative} & CVPR'22 & - & - & - & - & 42.2 \\
         & BAM~\cite{lang2022learning} & CVPR'22 & 43.4 & 50.6 & 47.5 & 43.4 & 46.2 \\
         & MSANet~\cite{iqbal2022msanet} & - & \textbf{47.8} & \textbf{57.4} &  \textbf{48.6} & \textbf{50.4} & \textbf{51.1}\\
    \midrule
        \multirow{3}{*}{Image Labels} & PANet~\cite{wang2019panet} & ICCV'19 & 12.7 & 8.7 &  5.9 & 4.8 & 8.0\\
        & Pix-MetaNet~\cite{lee2022pixel} & WACV'22 & 24.2 & 12.9 & \textbf{17.0} & 14.0 & 17.0 \\
         & Ours (WLSegNet)& - & \textbf{34.9} &\textbf {23.4} & 12.4 & \textbf{18.3 }& \textbf{22.2}\\
    \bottomrule
    \end{tabular}
    }
    
    \label{tab:wfss_coco}
\end{table}

\begin{table}[!h]
    \begin{minipage}{.5\linewidth}
   \centering
     \caption{2-way 1-shot FSS on Pascal-5$^i$.}
    \setlength{\tabcolsep}{4pt}
    \scalebox{0.6}{
    \begin{tabular}{clccccc}
    \toprule
        \multirow{2}{*}{Sup} & \multirow{2}{*}{Method} & \multicolumn{4}{c}{Fold mIOU} & \multirow{2}{*}{mean}\\
        & & 0 & 1 & 2 & 3 & \\
    \midrule
        \multirow{2}{*}{Pix}  & Pix-MetaNet & 36.5 & 51.8   & 48.5   & 38.9 & 43.9  \\
        & PANet & - & - & - & -  & \textbf{45.1} \\
    \midrule
        \multirow{3}{*}{Img} & PANet & 24.5  & 33.6 & 26.3  & 20.3 & 26.2  \\
        & Pix-MetaNet & 31.5 & 46.7  & 41.4  & 31.2 & 37.7  \\
         & Ours (WLSegNet) & \textbf{50.9} & \textbf{52.9} & \textbf{45.5} & \textbf{53.4} & \textbf{50.7} \\
    \bottomrule
    \end{tabular}
    }
    \label{tab:2way_pascal}
    \end{minipage}%
    \begin{minipage}{.5\linewidth}
    \centering
     \caption{2-way 1-shot FSS on COCO-20$^i$.}
    \setlength{\tabcolsep}{4pt}
    \scalebox{0.6}{
    \begin{tabular}{clccccc}
    \toprule
        \multirow{2}{*}{Sup} & \multirow{2}{*}{Method} & \multicolumn{4}{c}{Fold mIOU} & \multirow{2}{*}{mean}\\
        & & 0 & 1 & 2 & 3 & \\
    \midrule
        \multirow{1}{*}{Pix}  & Pix-MetaNet & 18.2 & 12.2   & 9.1  & 6.5 & 11.5  \\
    \midrule
        \multirow{2}{*}{Img} & Pix-MetaNet & 17.4 & 9.5  & 10.4  & 7.1 & 11.1 \\
         & Ours (WLSegNet) & \textbf{38.0} & \textbf{33.8} & \textbf{29.4} & \textbf{31.4} & \textbf{33.1}\\
    \bottomrule
    \end{tabular}
    }
    \label{tab:2way_coco}
    \end{minipage}
\end{table}

\begin{figure}[!h]
\centering
\includegraphics[width=1\linewidth]{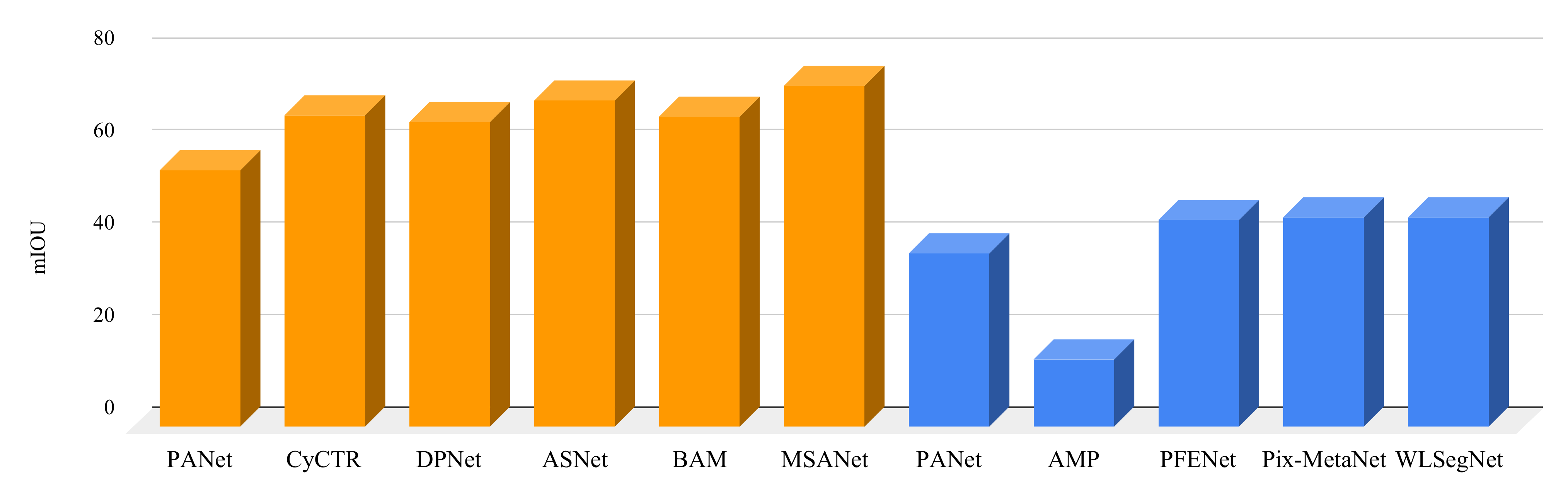}
\caption{1-way 5-shot FSS on Pascal-5$^i$. Orange bars represent strong supervision while blue bars represent weak supervision via image labels.}
\label{fig:1way5shotpascal}
\end{figure}

\begin{figure}[!h]
\centering
\includegraphics[width=1\linewidth]{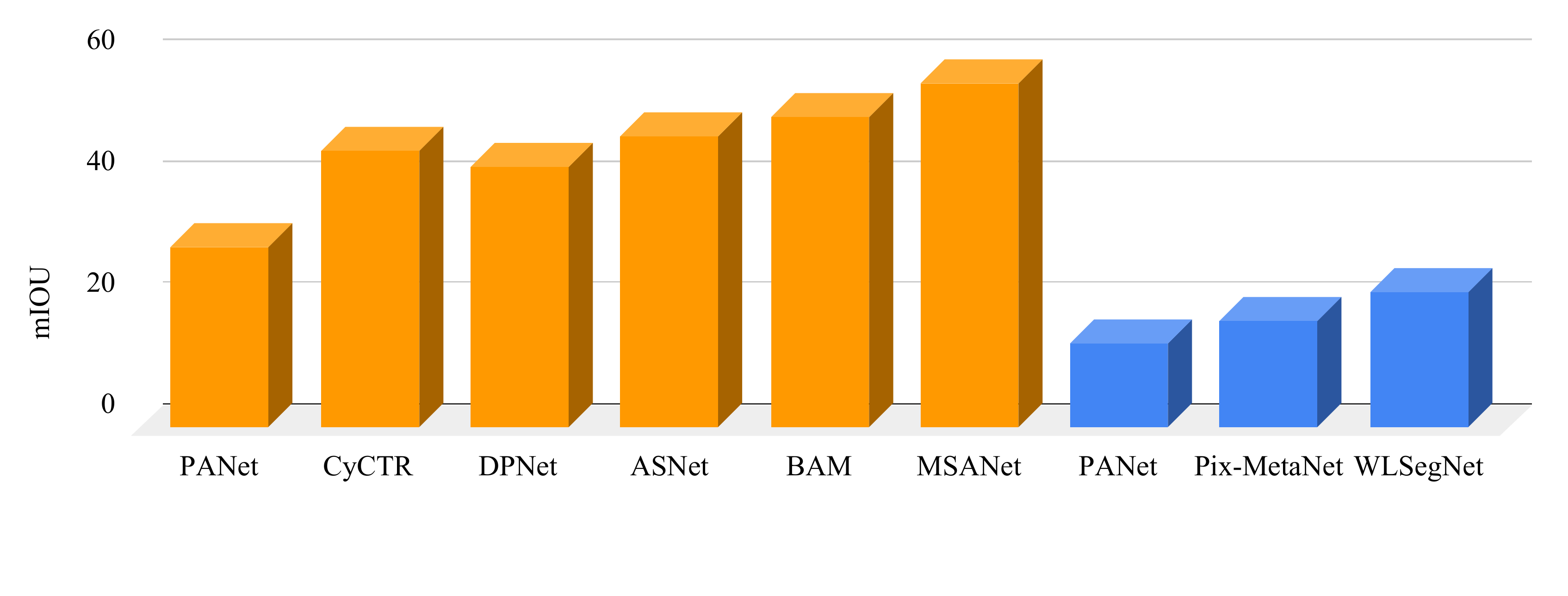}
\caption{1-way 5-shot FSS on COCO-20$^i$. Orange bars represent strong supervision while blue bars represent weak supervision via image labels.}
\label{fig:1way5shotcoco}
\end{figure}

\begin{figure}[!h]
\begin{minipage}{0.5\textwidth}
\centering
\includegraphics[width=1.0\linewidth]{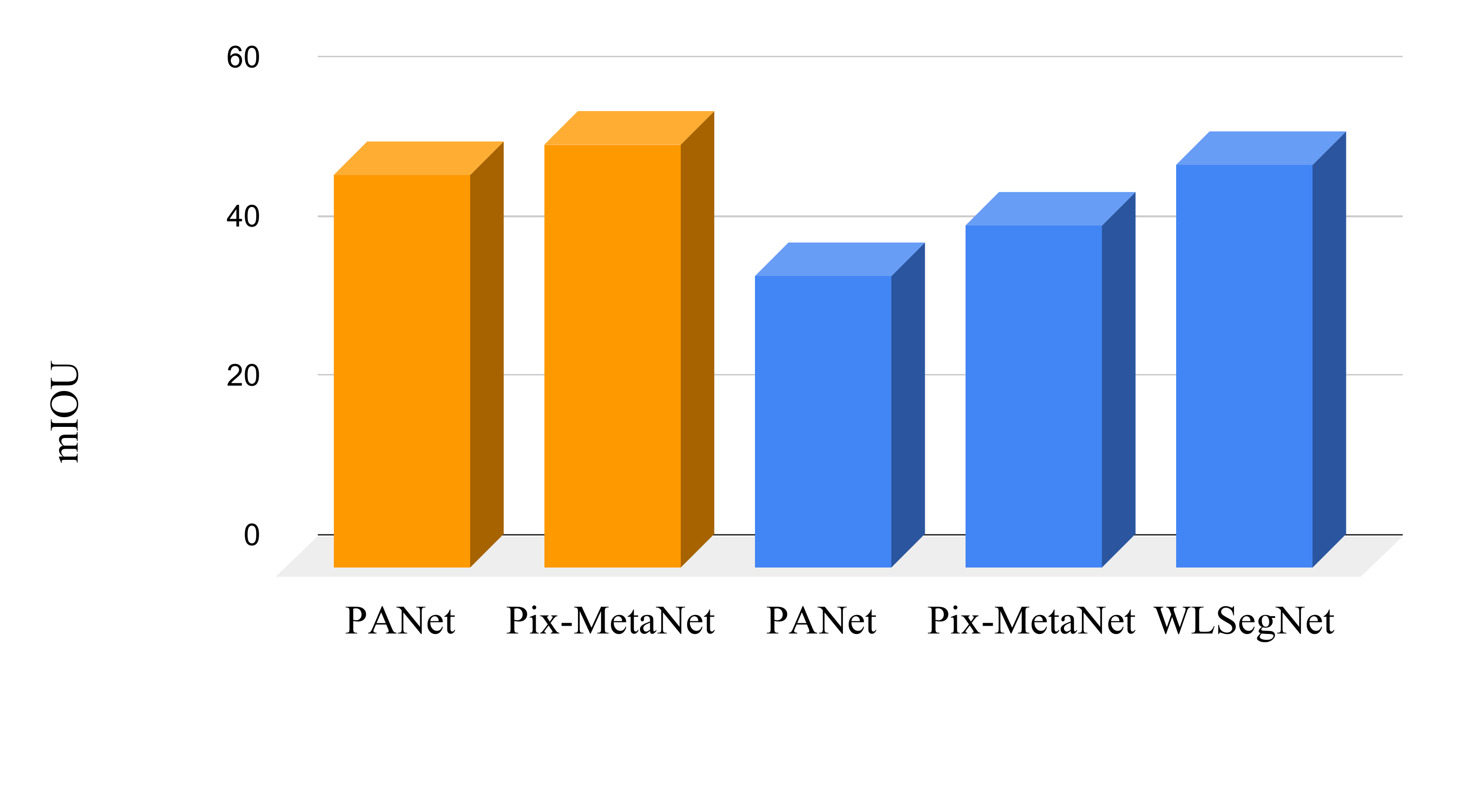}
\end{minipage}%
\begin{minipage}{0.5\textwidth}
\centering
\includegraphics[width=0.95\linewidth]{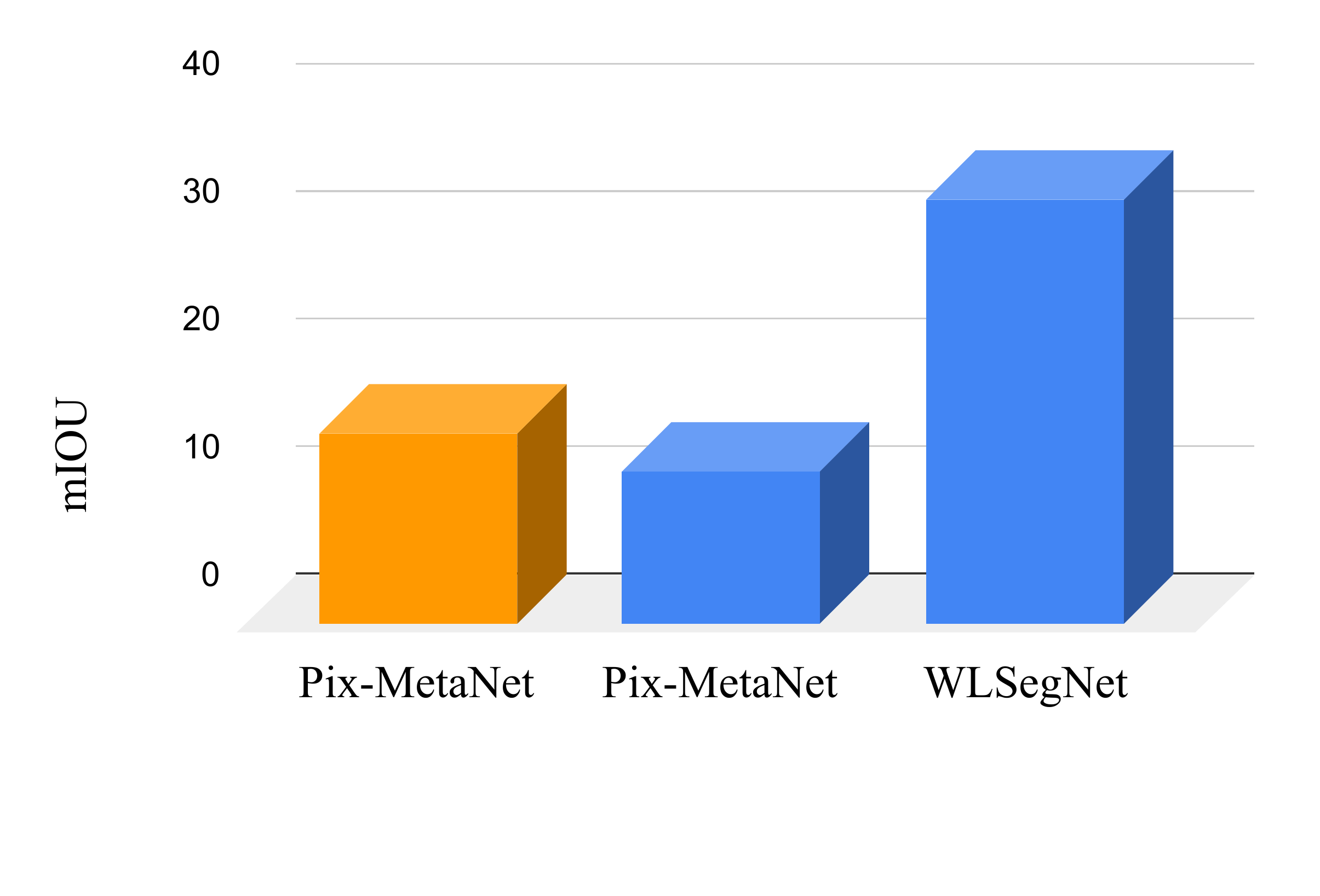}
\end{minipage}
\caption{2-way 5-shot FSS on Pascal-5$^i$ (left) and COCO-20$^i$ (right). Orange bars represent strong supervision while blue bars represent weak supervision via image labels.}
\label{fig:2way5shotPC}
\end{figure}

\subsubsection{Weakly Supervised Few-Shot Segmentation}
The performance of our approach in WFSS can be seen in Table~\ref{tab:wfss_pascal} and Table~\ref{tab:wfss_coco}. As evident from the results, we beat all methods using weak supervision by at least 7\% mIOU on PASCAL VOC and at least 30\% mIOU on COCO. Besides the commonly used 1-way 1-shot setting, we also experiment with 2-way 1-shot FSS in Table~\ref{tab:2way_pascal} and Table~\ref{tab:2way_coco}. Again, we beat weakly supervised baselines by huge margins. Our performance here exceeds all baselines by at least 13 and 22 mIOU points for PASCAL VOC and COCO respectively. On \{1,2\}-way 5-shot setting, WLSegNet clearly outperforms image-level baselines while being a strong contender for methods availing pixel-level supervision as observed from Figure~\ref{fig:1way5shotpascal} to Figure~\ref{fig:2way5shotPC}. 



\subsubsection{Cross-dataset Segmentation}
Following the setting of~\cite{Boudiaf_2021_CVPR}, we evaluate the performance of WLSegNet on the novel classes of PASCAL VOC with the COCO-trained model \textit{without fine-tuning}. These experiments test the ability of WLSegNet to handle domain shift between the classes of the two different datasets in the WZSS and WFSS settings. The novel PASCAL VOC classes are shown in Table~\ref{tab:cross-dataset}. The categories in fold $i$ are the novel classes in PASCAL
VOC after removing the seen classes in the corresponding training split on fold $i$ of COCO-20$^i$. We benchmark the performance of WLSegNet on the Cross-dataset setting in Table~\ref{tab:cross_data_res}. It is clearly evident from the results that even with domain shift, the generalizable prompts learned with WLSegNet help to deliver performance competitive with the pixel-based methods.

\begin{table}[!h]
    \centering
    \setlength{\tabcolsep}{8pt}
    \caption{Novel classes in each fold of the PASCAL VOC dataset in the Cross-dataset segmentation setting.}
    \scalebox{0.6}{
    \begin{tabularx}{\linewidth}{LLLL}
    \toprule
        fold 0 & fold 1 & fold 2 & fold 3 \\
    \midrule
        aeroplane, boat, chair,
diningtable, dog, person & bicycle, bus,
horse, sofa & bird, car, pottedplant,
sheep, train, tvmonitor & bottle, cat,
cow, motorbike  \\
    \bottomrule
    
    \end{tabularx}}
    \label{tab:cross-dataset}
\end{table}

\begin{table}[!h]
\centering
\caption{
ZSS and FSS on novel classes of PASCAL VOC when the model is trained on COCO-20$^i$. The second best results are underlined.
}
\scalebox{0.6}
{
    \centering
    \begin{tabular}{clcccccc}
    \toprule
        \multirow{2}{*}{Supervision} & \multirow{2}{*}{Method} & \multirow{2}{*}{Setting} & \multicolumn{4}{c}{Fold mIOU} & \multirow{2}{*}{mean}\\
        & & & 20$^0$ & 20$^1$ & 20$^2$ & 20$^3$ & \\
    \midrule
        \multirow{2}{*}{Pixel Labels} & LSeg~\cite{li2022languagedriven} &  zero-shot & \underline{24.6} & - & \underline{34.7} & 35.9 & 31.7\\
         & Fusioner~\cite{ma2022open} &  zero-shot & \textbf{39.9}  & \textbf{70.7} & \textbf{47.8} & \textbf{67.6} & \textbf{56.5}\\
         \midrule
         \multirow{1}{*}{Image Labels} & Ours (WLSegNet) & weak zero-shot & 17.6  & \underline{50.3} & 19.5 & \underline{52.4} & \underline{34.9} \\
         \midrule
          \multirow{3}{*}{Pixel Labels} & RPMM~\cite{10.1007/978-3-030-58598-3_45} &  1-way 1-shot & 36.3 & 55.0 & 52.5 & 54.6 & 49.6\\
         & PFENet~\cite{9154595} &  1-way 1-shot & 43.2 & \textbf{65.1} & \textbf{66.5} & \textbf{69.7} & \textbf{61.1}\\
         & CWT~\cite{Lu_2021_ICCV} &  1-way 1-shot & \textbf{53.5}  & \underline{59.2} & \underline{60.2} & \underline{64.9} & \underline{59.5}\\
    \midrule
        \multirow{1}{*}{Image Labels} & Ours (WLSegNet) & weak 1-way 1-shot & \underline{44.1} & 44.2 & 37.1 & 60.3 & 46.4 \\
    \bottomrule
    \end{tabular}
    }
    \label{tab:cross_data_res}
\end{table}

\begin{figure*}[!h]
\centering
\includegraphics[width=0.7\linewidth]{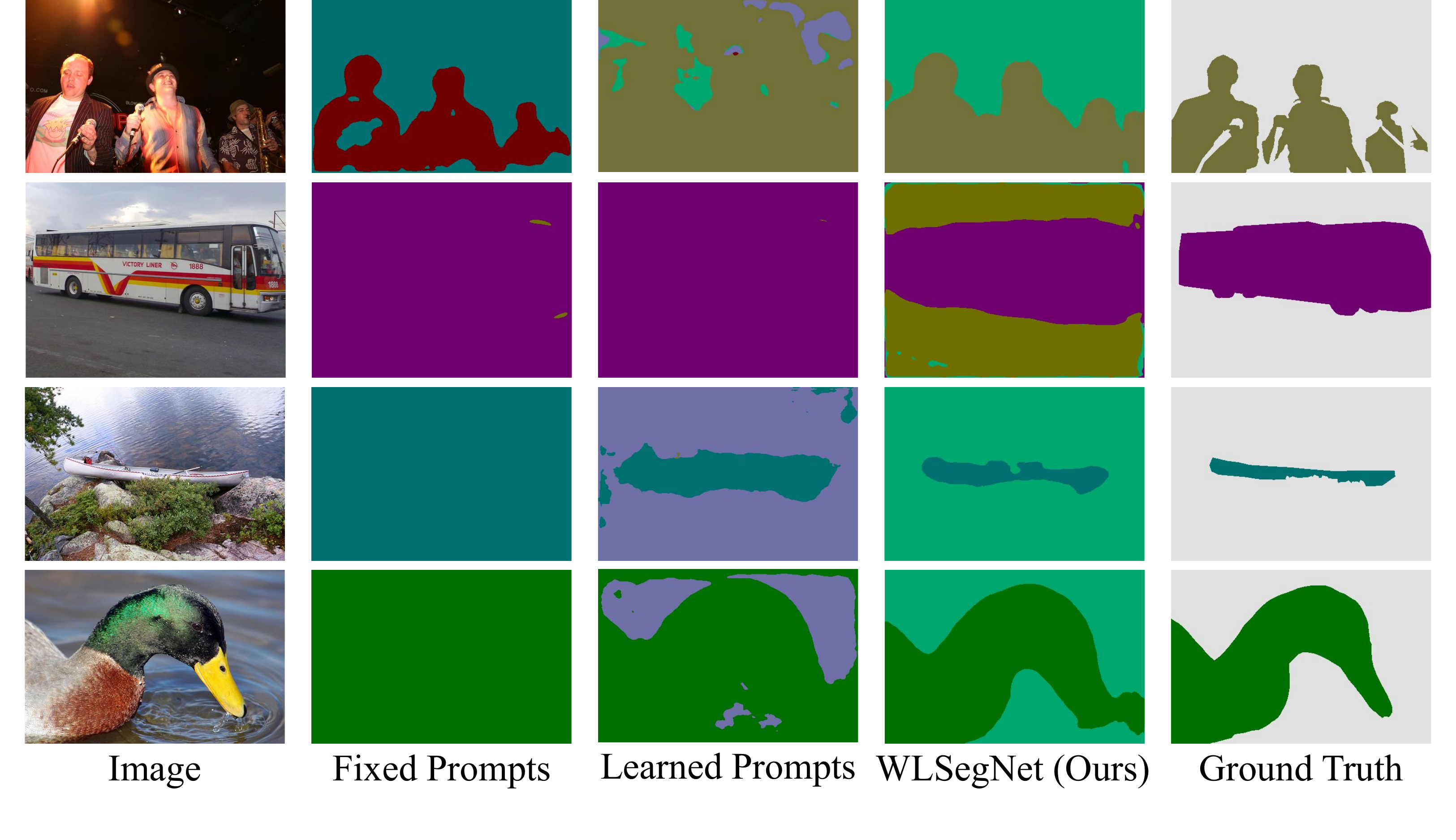}
\caption{Predicted masks with different prompt learning strategies in Weak Generalized Zero-Shot Segmentation (WGZSS) setting on PASCAL VOC.}
\label{fig:vis_wzss}
\end{figure*}

\begin{figure*}[!h]
\centering
\includegraphics[width=0.72\linewidth]{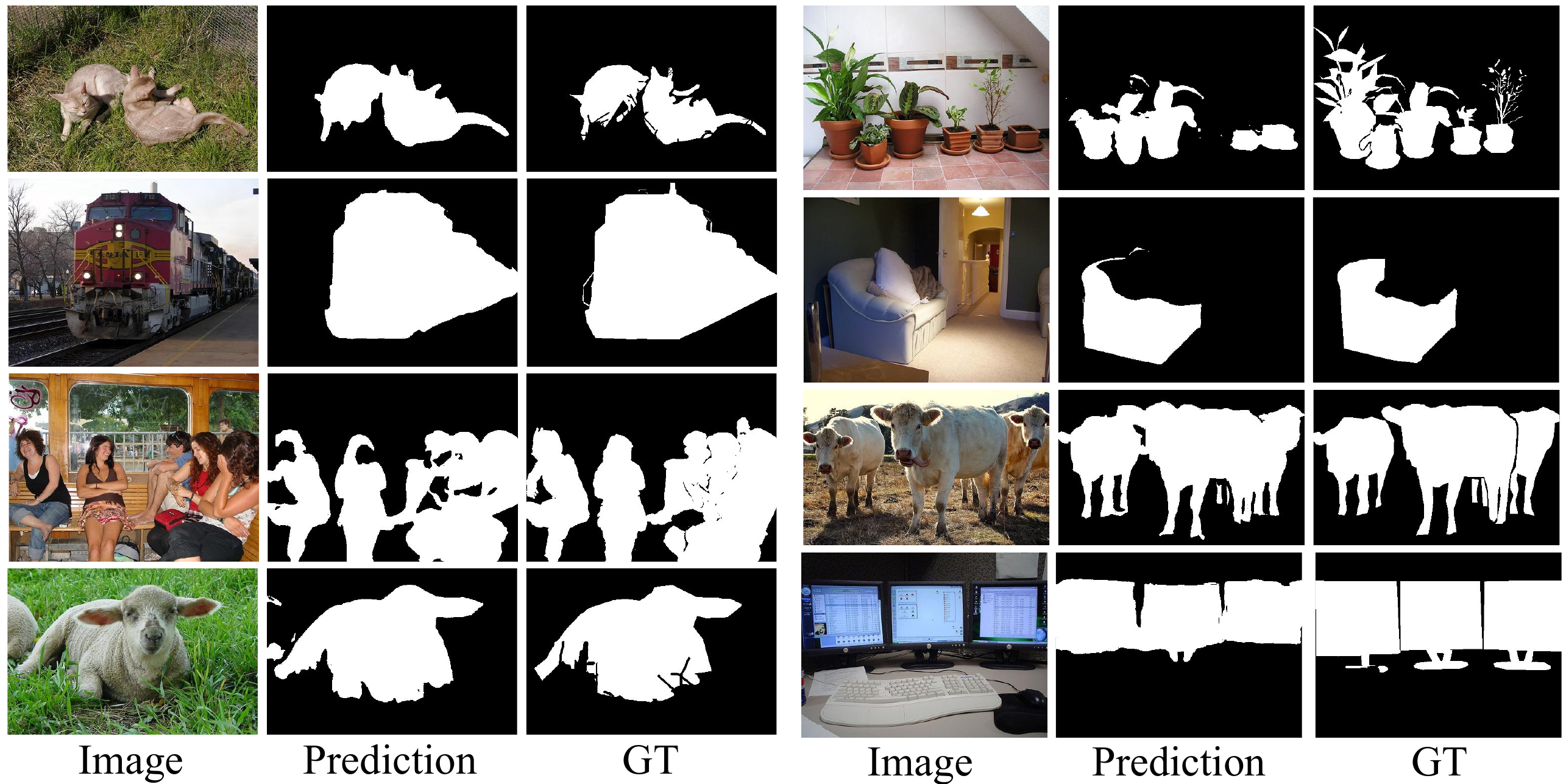}
\caption{1-way 1-shot predicted masks on Pascal-5$^i$.}
\label{fig:1way_vis_pascal}
\end{figure*}

\subsubsection{Qualitative Analysis}
\begin{figure*}[!h]
\centering
\includegraphics[width=0.72\linewidth]{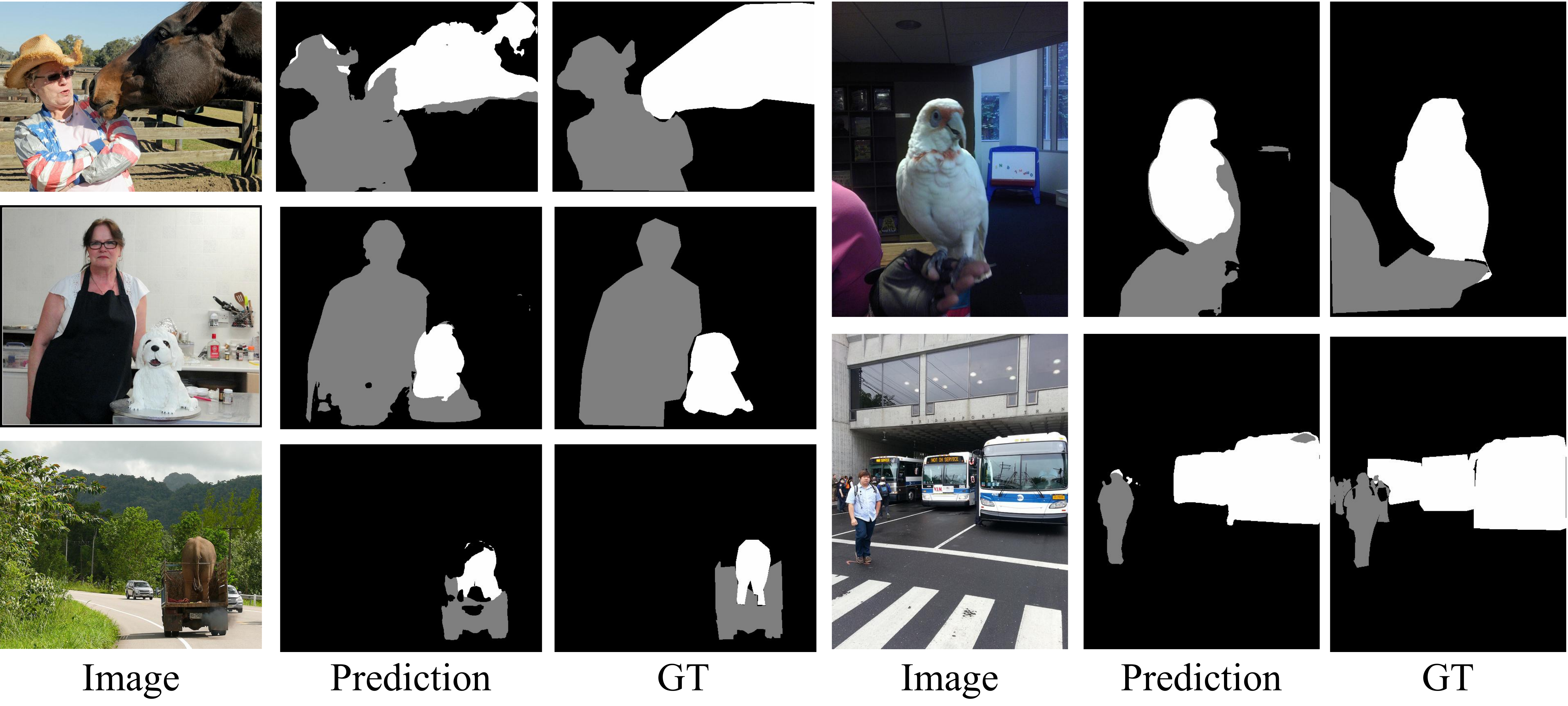}
\caption{2-way 1-shot predicted masks on COCO-20$^i$.}
\label{fig:2way_vis}
\end{figure*}

We visualize the predicted masks in different settings by WLSegNet which gives compelling results in the weak Few-Shot and Zero-Shot Segmentation. As observed from Figure~\ref{fig:vis_wzss}, the proposed prompt learning strategy is able to capture complex objects while other strategies fail to segment the desired seen/unseen classes in the weak Generalized Zero-Shot Segmentation (WGZSS) setting. Similarly, in a comparatively harder setting of 2-way (Figure~\ref{fig:2way_vis}) with a large-scale dataset like COCO, WLSegNet is able to segment the required target classes having different sizes thereby closely matching the Ground Truth (GT).

\begin{figure*}[!h]
\centering
\includegraphics[width=0.72\linewidth]{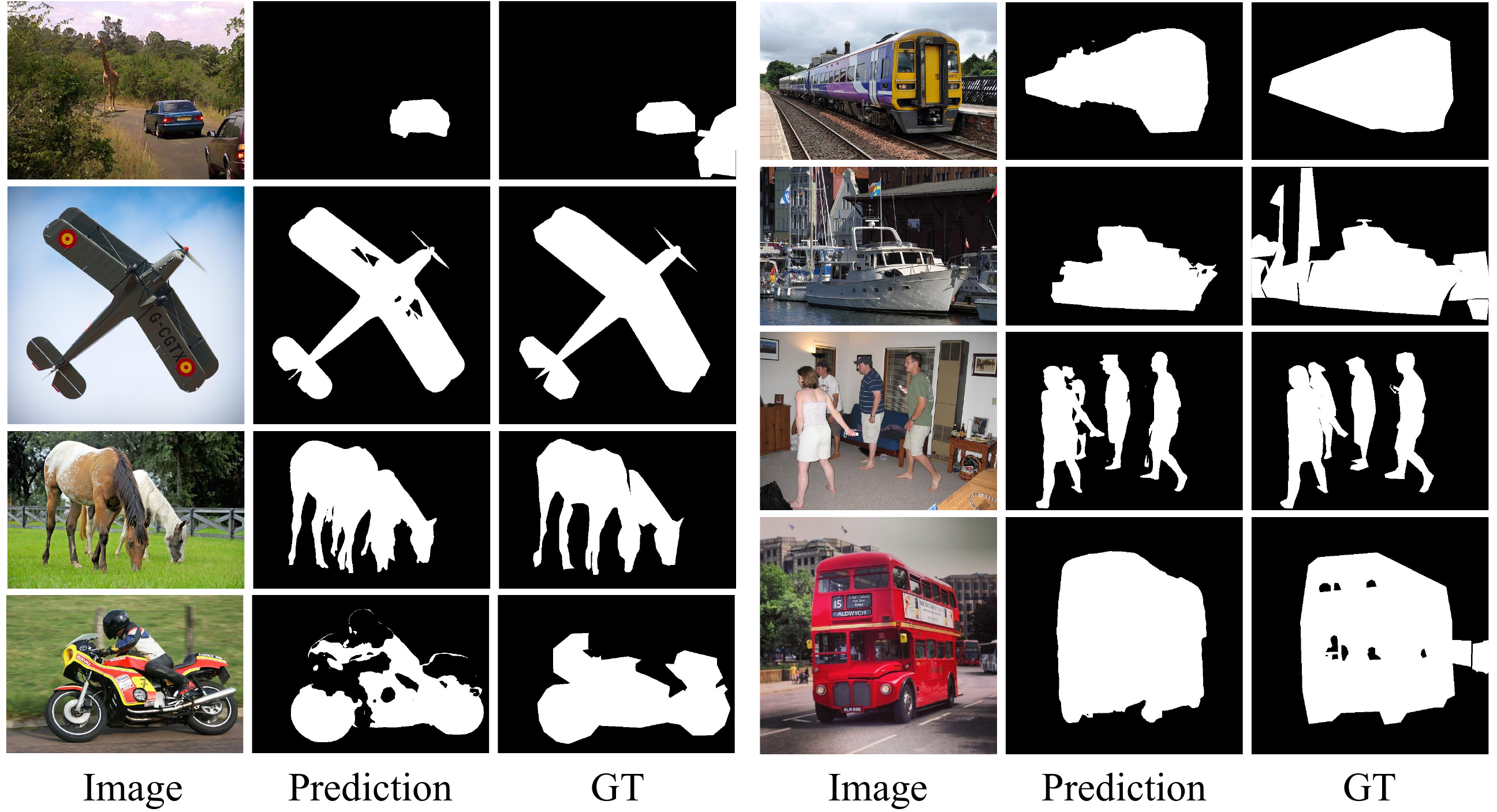}
\caption{1-way 1-shot predicted masks on COCO-20$^i$.}
\label{fig:1way_vis}
\end{figure*}

\begin{figure}[!h]
\begin{minipage}{0.5\textwidth}
\centering
\includegraphics[width=1\linewidth]{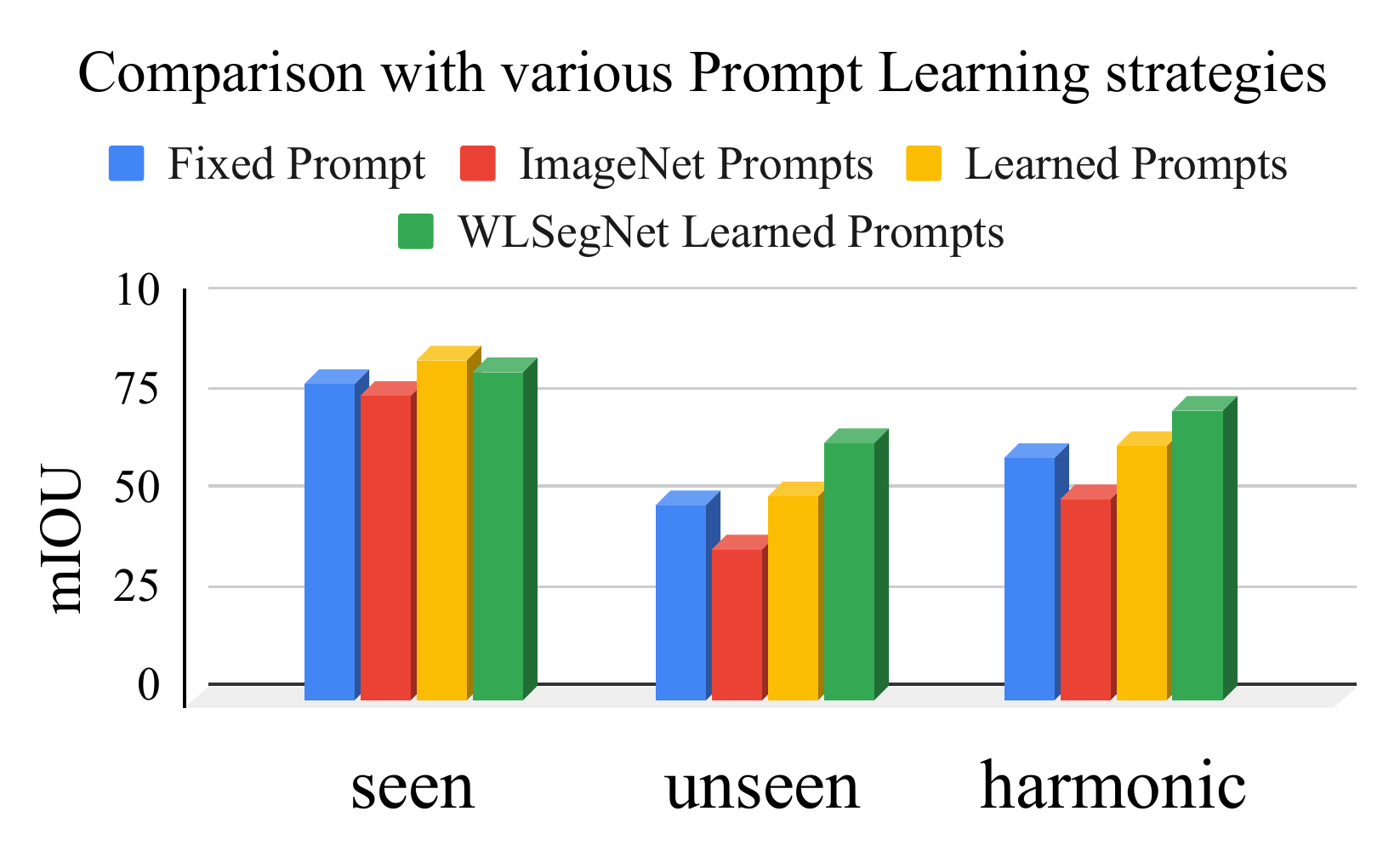}
\end{minipage}%
\begin{minipage}{0.5\textwidth}
\centering
\includegraphics[width=0.95\linewidth]{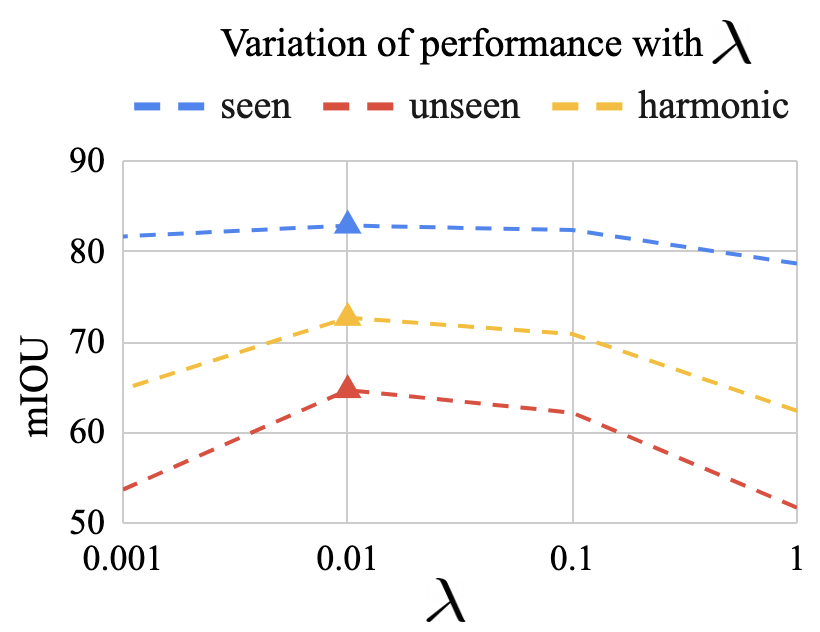}
\end{minipage}
\caption{Ablation with different prompting strategies.}
\label{fig:ablation}
\end{figure} 

\subsection{Ablation Studies}

We perform further experiments to understand the relative contributions of the various components in our approach. First, we experiment with different strategies to get text prompts. We try fixed prompts, where a single prompt template is used for all classes; ImageNet prompts, where a prompt template is chosen randomly from 80 prompts designed for ImageNet; Learned prompts, similar to the ones used in \cite{xu2021simple} and finally ours. Figure~\ref{fig:ablation} (left) shows that our prompt learning method performs better for the unseen classes resulting in the highest harmonic mIOU. In Figure~\ref{fig:ablation} (right), we analyse the performance for different values of the hyperparameter $\lambda$ as used in Eq~(\ref{eq:lambda}). 

\begin{figure}[!h]
\centering
\includegraphics[width=0.49\linewidth]{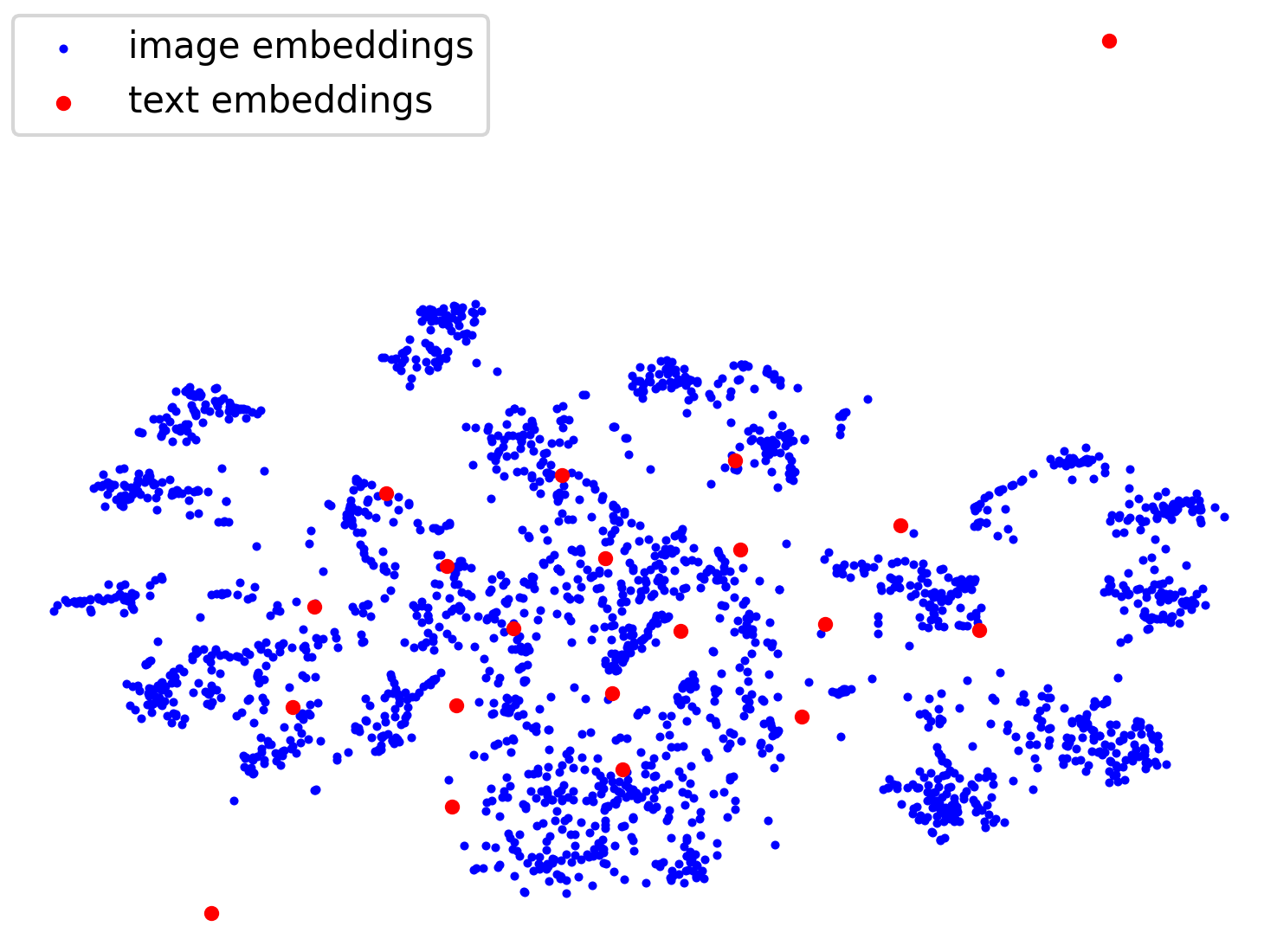}
\includegraphics[width=0.49\linewidth]{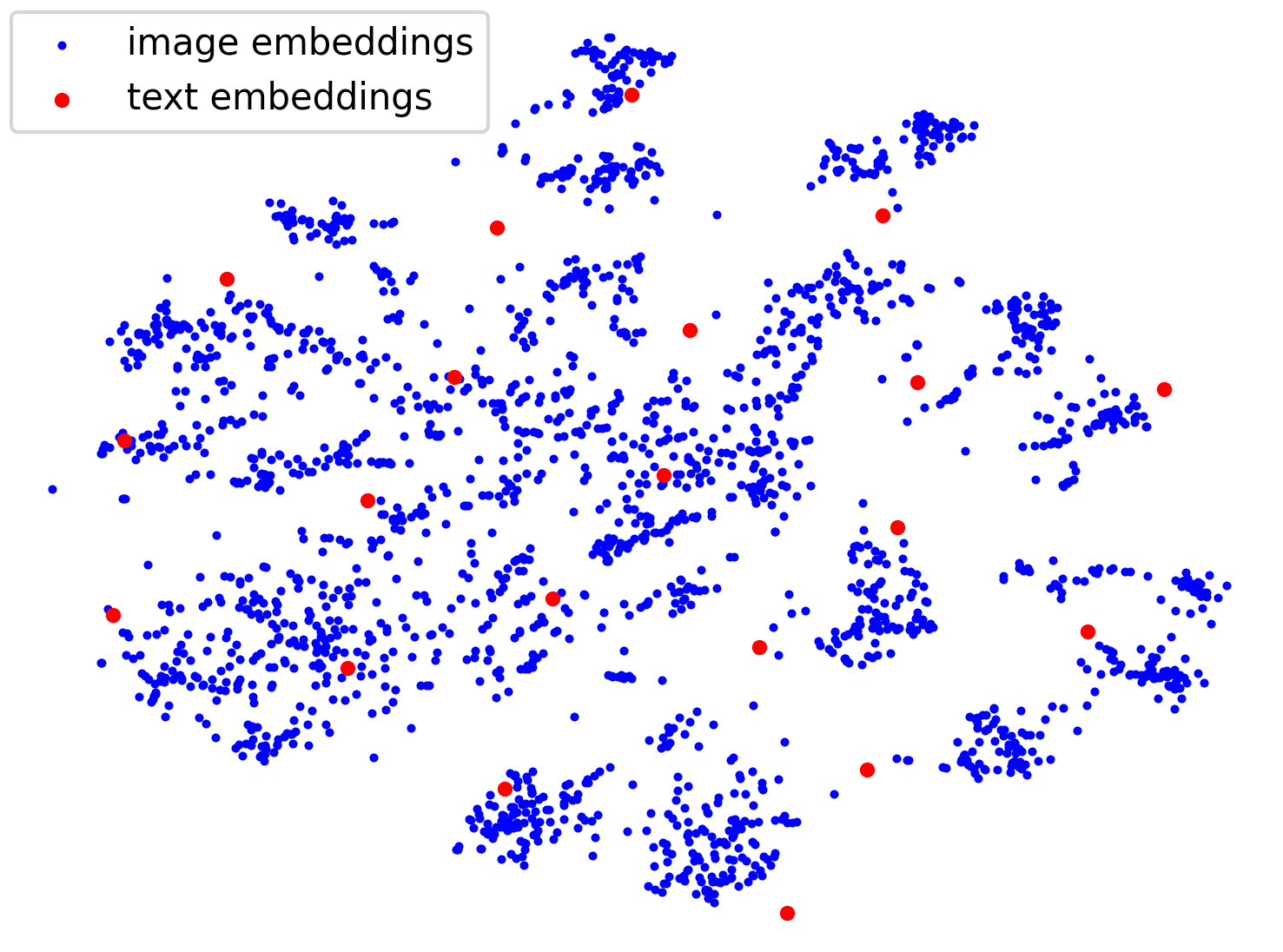}
\caption{t-SNE plots showing image and text features obtained from different prompt learning methods. The left figure is obtained using SimSeg~\cite{xu2021simple} and the right from WLSegNet.}
\label{fig:tsne}
\end{figure}

While experimenting with different batch sizes, we observe that the method is not sensitive to changes here. We evaluate the performance of WLSegNet (Table~\ref{tab:vary_camg_clip_plg}) by varying the mask proposal generation methods in the CAMG module, CLIP backbones and pseudo-label generation methods in the PLG module. These experiments helped to design and optimize the CAMG and PLG modules and in the selection of the backbone architecture for CLIP. Finally, we visualize the features of images obtained from the image encoder and the text features of prompts of different classes in Figure~\ref{fig:tsne}. The image features do roughly form clusters in the feature space and red text features corresponding to the classes are roughly aligned with the centres of these clusters. The clusters are better formed and text features are better aligned in the plot on the right, further demonstrating the generalizability of the learned prompts. All these ablation studies are performed on the PASCAL VOC dataset in the WGZSS setting. 

\begin{table}[!h]
    \centering
    \caption{WLSegNet performance (harmonic mIOU) with different mask  proposal generation methods for the Class-Agnostic Mask Generation (CAMG) module,  different CLIP backbones and different pseudo-label generation methods for the Pseudo Label Generation (PLG) module in the weak Generalized Zero-Shot Segmentation (WGZSS) setting on the PASCAL VOC dataset.}
    \scalebox{0.6}{
    \begin{tabular}{cc|cc|cc}
    \toprule
        CAMG & harmonic mIOU & CLIP backbone & harmonic mIOU  & PLG & harmonic mIOU  \\
    \midrule
        GPB-UCM~\cite{arbelaez2010contour} & 36.3 & ResNet50 & 58.8 & RCA~\cite{zhou2022regional} & 68.7\\
        Selective Search~\cite{uijlings2013selective} & 36.6 & ResNet101 & 53.4 & L2G~\cite{jiang2022l2g} & 70.8\\
        MaskFormer~\cite{cheng2021per} & 70.8 & ViT-B/16 & 70.8 & - & -\\
    \bottomrule 
    \end{tabular}
    }
    \label{tab:vary_camg_clip_plg}
\end{table}

\section{Conclusion}

Data-efficient problem settings like Open Vocabulary Semantic Segmentation (OVSS) are of utmost importance for an intelligent model because of the similar difficulties existing in many real-world scenarios. Extensive research is being done to develop novel methods that require significantly lesser annotation costs while maintaining expected standards of performance. We explore one such challenging domain (OVSS) where a model is expected to generalize to a wide range of classes it never sees during training while also only relying on relatively inexpensive weak annotations and vision-language models like CLIP. In a unified approach to weakly supervised Zero and Few-Shot segmentation, we overcome certain limitations reported by existing works and learn a label-efficient model and prompts that are highly generalizable to unseen classes. The superior performance of our method is corroborated by extensive experimentation on two large-scale datasets. We hope this work will promote further research in this relatively under-explored domain and provide a strong baseline to benchmark new methods.

 \bibliographystyle{elsarticle-num} 
 \bibliography{cas-refs}

\end{document}